\documentclass[journal,twoside]{IEEEtran}
\usepackage{cite}

\ifCLASSINFOpdf
  \usepackage[pdftex]{graphicx}
\else
\fi
\usepackage{amsmath}
\usepackage{algorithmic}

\usepackage{array}

\ifCLASSOPTIONcompsoc
  \usepackage[caption=false,font=normalsize,labelfont=sf,textfont=sf]{subfig}
\else
  \usepackage[caption=false,font=footnotesize]{subfig}
\fi
\usepackage{fixltx2e}

\usepackage{stfloats}
\usepackage{url}

\usepackage{booktabs} 
\usepackage{multirow}
\usepackage{xcolor}
\usepackage[super]{nth}
\usepackage{tikz}
\newcommand*\circled[1]{\tikz[baseline=(char.base)]{
		\node[shape=circle,draw,inner sep=0.2pt] (char) {#1};}}
\usepackage{soul}
\usepackage{enumitem}
\usepackage{algorithm}

\usepackage[none]{hyphenat}

\begin{document}
%
\title{FSpiNN: An Optimization Framework for Memory- and Energy-Efficient Spiking Neural Networks\\}

\author{Rachmad~Vidya~Wicaksana~Putra and Muhammad~Shafique,~\IEEEmembership{Senior Member, IEEE}
\thanks{Manuscript received April 18, 2020; revised June 12, 2020; accepted July 6, 2020. This article was presented in the International Conference on Compilers, Architecture, and Synthesis for Embedded Systems 2020 and appears as part of the ESWEEK-TCAD special issue.}
\thanks{R. V. W. Putra is with Technische Universit\"at Wien (TU Wien), Vienna, Austria. E-mail: rachmad.putra@tuwien.ac.at}
\thanks{M. Shafique is with Division of Engineering, New York University Abu Dhabi (NYU AD), Abu Dhabi, United Arab Emirates, and Institute of Computer Engineering, Technische Universität Wien (TU Wien), Vienna, Austria. E-mail: muhammad.shafique@nyu.edu and muhammad.shafique@tuwien.ac.at}
\vspace{-0.3cm}}

%
%

\markboth{To appear at the IEEE Trans. on Computer-Aided Design of Integrated Circuits and Systems 2020 (ESWEEK-TCAD Special Issue)}%
{Putra \MakeLowercase{\textit{et al.}}: FSPINN: An Optimization Framework for Memory- and Energy-Efficient SNNs}%
%



\maketitle

\begin{abstract}
Spiking Neural Networks (SNNs) are gaining interest due to their event-driven processing which potentially consumes low power/energy computations in hardware platforms, while offering unsupervised learning capability due to the spike-timing-dependent plasticity (STDP) rule. 
However, state-of-the-art SNNs require a large memory footprint to achieve high accuracy, thereby making them difficult to be deployed on embedded systems, for instance on battery-powered mobile devices and IoT Edge nodes. 
Towards this, we propose FSpiNN, an optimization framework for obtaining memory- and energy-efficient SNNs for training and inference processing, with unsupervised learning capability while maintaining accuracy. 
It is achieved by (1) reducing the computational requirements of neuronal and STDP operations, (2) improving the accuracy of STDP-based learning, (3) compressing the SNN through a fixed-point quantization, and (4) incorporating the memory and energy requirements in the optimization process. 
FSpiNN reduces the computational requirements by reducing the number of neuronal operations, the STDP-based synaptic weight updates, and the STDP complexity. 
To improve the accuracy of learning, FSpiNN employs timestep-based synaptic weight updates, and adaptively determines the STDP potentiation factor and the effective inhibition strength.
The experimental results show that, as compared to the state-of-the-art work, FSpiNN achieves 7.5x memory saving, and improves the energy-efficiency by 3.5x on average for training and by 1.8x on average for inference, across MNIST and Fashion MNIST datasets, with no accuracy loss for a network with 4900 excitatory neurons, thereby enabling energy-efficient SNNs for edge devices/embedded systems.  
\end{abstract}


%
\IEEEpeerreviewmaketitle

\section{Introduction}
\label{Sec:Intro}

\IEEEPARstart{T}{he} spiking neural networks (SNNs) are rapidly gaining research interest since they have shown great potential in completing various machine learning tasks, while exhibiting high biological plausibility \cite{Ref:Pfeiffer_DLSNN_FNINS18, Ref:Tavanaei_DLSNN_Neunet18, Ref:Shafique_DnT20_RobustML, Ref:Marchisio_arXiv19_SpikeSecure, Ref:Venceslai_arXiv20_NeuroAttack}. 
That is, the SNNs mimic the behavior of biological spiking networks through (i) event-driven processing, and (ii) spike-timing-dependent plasticity (STDP)-based unsupervised learning. 
The event-driven processing potentially enables low power/energy computation in the neuromorphic hardware, such as \cite{ Ref:Davies_Loihi_MM18}\cite{Ref:Massa_arXiv20_Loihi}, due to its sparse spiking-based computation. 
The STDP-based learning enables SNNs to learn information from the unlabeled data, which is desired for real-world applications, as gathering unlabeled data is easier and cheaper than labeled data \cite{Ref:Rathi_PruneQuantizeSNN_TCAD18}. 
Thus, SNNs bear the potential to obtain better algorithmic performance (e.g., classification accuracy) with lower power/energy consumption than other neural network algorithms in the unsupervised-learning settings \cite{Ref:Pfeiffer_DLSNN_FNINS18}.

\vspace{-0.3cm}
\subsection{Targeted Research Problem}
\label{Sec:ResearchProblem}

\begin{figure}[hbtp]
\vspace{-0.3cm}
\centering
\includegraphics[width=\linewidth]{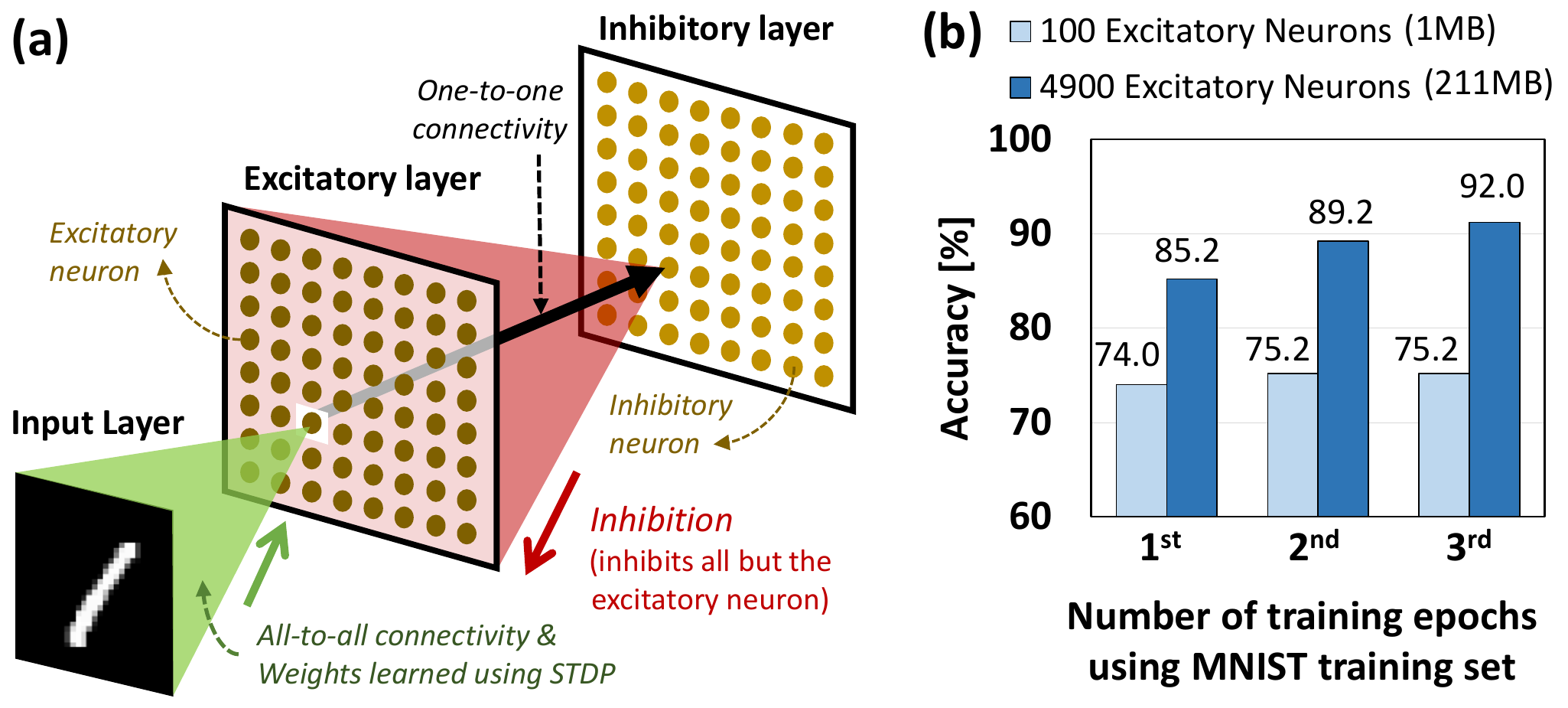}
\vspace{-0.5cm}
\caption{(a) An SNN architecture considered in this work is from \cite{Ref:Diehl_STDPmnist_FNCOM15}\cite{Ref:Srinivasan_EnhPlast_IJCNN17}. 
(b)~A~large-sized SNN typically achieves higher classification accuracy, e.g., the accuracy of an 1MB-sized SNN with a total of 200 neurons (100 excitatory and inhibitory neurons) is lower than a 211MB-sized SNN with a total of 9800 neurons (4900 excitatory and inhibitory neurons) on MNIST \cite{Ref:Lecun_MNIST_IEEE98}.}
\label{Fig:SNN_TopologyAccuracy}
\vspace{-0.2cm}
\end{figure}

In an SNN architecture with STDP \cite{Ref:Diehl_STDPmnist_FNCOM15} (depicted in Fig.~\ref{Fig:SNN_TopologyAccuracy}(a)), each \textit{excitatory} neuron is expected to recognize a class in the dataset, hence the connecting synapses from the same excitatory neuron have to learn the input features of a specific class (a detailed architecture discussion is provided in Section \ref{Sec:PrelimSNNs}). 
Previous works \cite{Ref:Hazan_SOMSNN_IJCNN18, Ref:Srinivasan_EnhPlast_IJCNN17, Ref:Panda_ASP_JETCAS18, Ref:Saunders_LCSNN_NeuNet19, Ref:Hazan_LMSNN_AMAI19} focus on improving the classification accuracy, but at the cost of a huge amount of additional computations, which leads to high energy and high memory footprint. 
For instance, the state-of-the-art work in \cite{Ref:Srinivasan_EnhPlast_IJCNN17} improves the effectiveness of the STDP-based learning by updating the weights every two or three postsynaptic spikes, to ensure that the update is essential.
This reduces the number of weight updates, but requires a 2-bit counter for each excitatory neuron to keep track of the number of postsynaptic spikes. 
Moreover, it needs a total of 200 neurons (100 excitatory and inhibitory neurons), to achieve $\sim$74\% accuracy in MNIST digit classification\footnote{Note: unlike deep neural networks (DNNs), the research for the unsupervised learning-based SNNs is still in early stage and mostly use small datasets like MNIST and Fashion MNIST. We adopt the same test conditions as used widely by the SNN research community \cite{Ref:Diehl_STDPmnist_FNCOM15, Ref:Hazan_SOMSNN_IJCNN18, Ref:Srinivasan_EnhPlast_IJCNN17, Ref:Panda_ASP_JETCAS18, Ref:Saunders_LCSNN_NeuNet19, Ref:Hazan_LMSNN_AMAI19, Ref:She_ParallelSpikeSim_DATE19}.}.
Although all these techniques result in an improvement in the classification accuracy, they incur high computational, energy, and memory costs. 
This is not desirable for embedded applications with a stringent constraints (for instance, in terms of computations, energy, and memory).

In the following, we present a motivational case study to illustrate the compute, memory, and communication requirements for an SNN executing on different hardware platforms, and highlight the associated research challenges.

\vspace{-0.2cm}
\subsection{Motivational Analysis and Associated Research Challenges}
\label{Sec:AnalysisChallenges}

\begin{figure}[hbtp]
\vspace{-0.1cm}
\centering
\includegraphics[width=\linewidth]{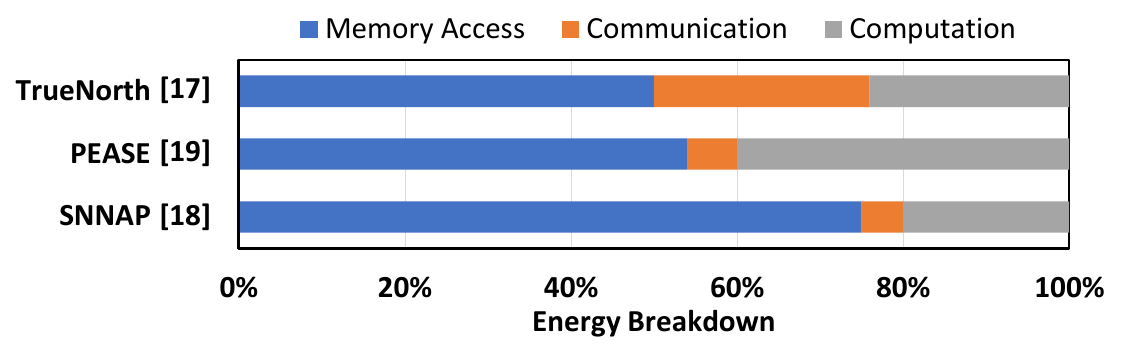}
\vspace{-0.5cm}
\caption{Energy breakdown of processing SNN in several hardware platforms (adapted from the studies presented in \cite{Ref:Krithivasan_SpikeBundle_ISLPED19}).}
\label{Fig:ObserveEnergyBreak}
\vspace{-0.1cm}
\end{figure}

In Fig.~\ref{Fig:SNN_TopologyAccuracy}(b), we observed that a large-sized SNN typically achieves higher classification accuracy and consumes a larger memory footprint. 
It shows that to achieve 92\% accuracy in MNIST digit classification, the SNN requires a total of 9800 neurons (4900 excitatory and inhibitory neurons) with 3 epochs of training, and consumes more than 200MB of memory.
On the other hand, most of the SNN hardwares employ a limited size of on-chip memory (e.g., less than 100MB) \cite{Ref:Akopyan_TrueNorth_TCAD15, Ref:Sen_ApproxSNN_DATE17, Ref:Roy_PEASE_ISLPED17, Ref:Frenkel_ODIN_TBCAS19}, which makes running a large-sized network (whose size is larger than the on-chip memory) energy-consuming. 
The reason is that this condition requires a high number of memory accesses, whose energy is typically higher than the compute operations \cite{Ref:Horowitz_ComputeEnergy_ISSCC14,Ref:Capra_MDPI20_DNNsurvey, Ref:Putra_arXiv20_DRMap}.
Previous work in \cite{Ref:Krithivasan_SpikeBundle_ISLPED19} observed that the memory accesses are dominant, consuming about 50\%-75\% energy of SNN processing in different hardware platforms \cite{Ref:Akopyan_TrueNorth_TCAD15, Ref:Sen_ApproxSNN_DATE17, Ref:Roy_PEASE_ISLPED17} (see Fig.~\ref{Fig:ObserveEnergyBreak}).

We also observed that there are inefficient computations that hinder SNNs to achieve higher energy-efficiency, that come from complex neuronal and STDP operations. 
They require exponential calculations for computing the membrane and threshold potential decay, and the synaptic trace and weight dependence, respectively (see details in Section \ref{Sec:Prelim}). 
Furthermore, there are ineffective STDP operations that come from spurious weight updates, which occur when the synapses of a neuron learn the overlapped features from different classes, thereby degrading the recognition capability of the neuron and also consuming energy. 
This happens since the general STDP rule updates the synaptic weight every pre- and post-synaptic spike (see details in Section \ref{Sec:Prelim}).  

\textbf{Required:} 
An optimization technique is required to reduce SNNs' memory and energy requirements for both training and inference processing, while maintaining the classification accuracy, thereby enabling their deployment on memory/energy-constrained embedded systems. 
However, developing such an optimization technique poses different design challenges as discussed below.

\textbf{Associated Research Challenges:}
The high memory requirements mainly come from a large number of parameters, such as synaptic weights and neuron parameters.
Reducing these parameters may degrade the classification accuracy.
Hence, the parameter reduction should be done by identifying and eliminating the non-significant parameters. 
Furthermore, bit-width quantization may also be employed, but it can also lead to accuracy degradation.
To overcome the limitations of the above optimization methods, the targeted research question is: if and how can we refine the STDP-based learning technique such that the classification accuracy is improved at minimal overhead. 

\vspace{-0.4cm}
\subsection{Our Novel Contributions}
\begin{figure}[hbtp]
\vspace{-0.1cm}
\centering
\includegraphics[width=\linewidth]{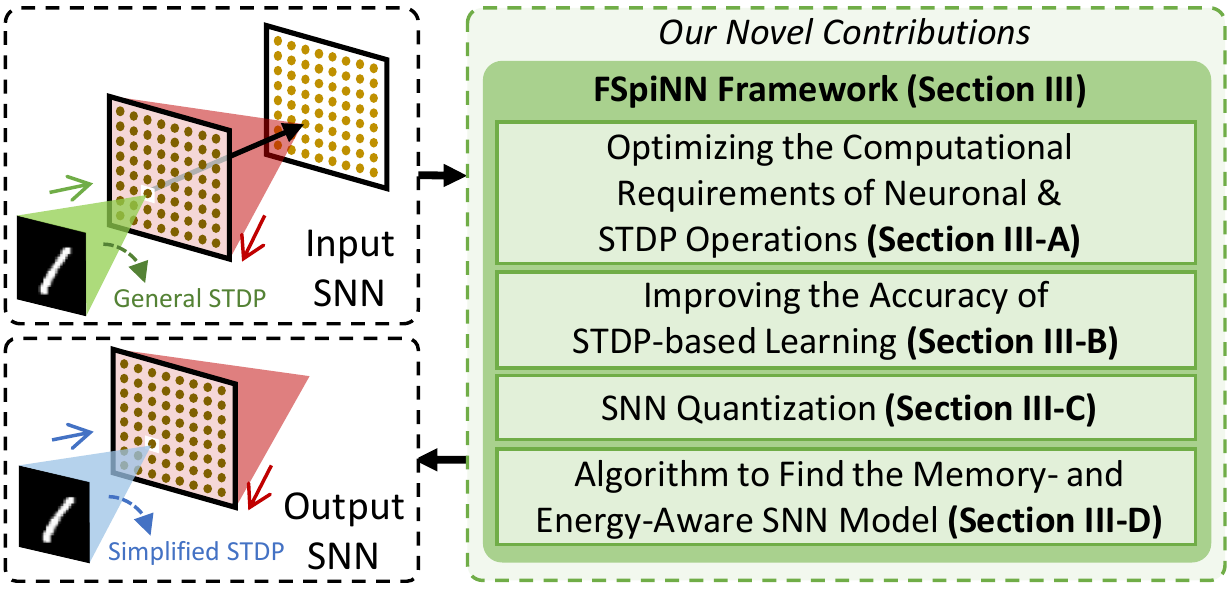}
\vspace{-0.5cm}
\caption{The overview of our novel contributions, which are highlighted in the green boxes.}
\label{Fig:NovelContributions}
\vspace{-0.1cm}
\end{figure}

To address the above challenges, we propose \textbf{FSpiNN}, a novel optimization framework for memory- and energy-efficient spiking neural networks for both training and inference, that employs the following key techniques (see an overview in Fig.~\ref{Fig:NovelContributions}) to overcome the above-discussed research challenges.

\begin{enumerate}[leftmargin=*]
    \item \textbf{Optimization of the neuronal and STDP  operations} by reducing (i) the inhibitory neurons through direct lateral inhibitory connections, (ii) the presynaptic spike-based weight updates, and (iii) the STDP complexity through elimination of the exponential calculation in the weight dependence part. 
    \item \textbf{An algorithm for improving the accuracy of STDP-based learning} by (i) minimizing the spurious weight updates through timestep-based operations, and (ii) effectively potentiating the weight in each update through an adaptive potentiation factor, and (iii) providing an effective competition among neurons through an adaptive inhibition. 
    \item \textbf{SNN quantization to compress the bit-width of network parameters:} It employs a fixed-point format by rounding to the nearest value, thereby providing a trade-off between the classification accuracy and the memory requirements.
    \item \textbf{An algorithm to find the memory- and energy-aware SNN model:} It incorporates the memory and energy requirements in the optimization process, and employs a search algorithm to find the desired model.
\end{enumerate}

\textbf{Key Results:} 
We evaluated our FSpiNN framework using a Python-based SNN simulator on an Nvidia GeForce GTX 1080 Ti GPU.
The experimental results show that, compared to the state-of-the-art \cite{Ref:Diehl_STDPmnist_FNCOM15}, our FSpiNN achieves 7.5x memory saving, and improves the energy-efficiency by 3.5x on average for training and by 1.8x on average for inference, across different datasets (MNIST and Fashion MNIST), with no accuracy loss for a network with 4900 excitatory neurons. 

\section{Preliminaries}
\label{Sec:Prelim}

\subsection{Spiking Neural Networks}
\label{Sec:PrelimSNNs}

Spiking Neural Networks (SNNs) are considered as the third generation of neural network computation models \cite{Ref:Maass_SNN_NeuNet97}, since they exhibit high biological plausibility. 
They mimic the behavior of biological spiking networks, i.e., action potentials or spikes are used to convey information. 
The SNN computational model is composed of spike/neural coding, network architecture, neuron model, synaptic model, and learning rule \cite{Ref:Mozafari_SpykeTorch_FNINS19}.

\textbf{Spike Coding:} It converts the information into a sequence of spikes (spike train).
Various spike coding methods have been studied in the literature, such as rate, temporal, rank-order, and phase coding schemes \cite{Ref:Gautrais_SpikeCoding_Bio98, Ref:Thorpe_RankOrder_Springer98, Ref:Kayser_PhaseCoding_Neuron09,Ref:Park_BurstSNN_DAC19}.
Here, we consider rate coding, since it has demonstrated high accuracy when employed in unsupervised learning-based SNNs.
Rate coding converts the intensity of a pixel to a spike train.
Typically, a higher intensity pixel is converted into a higher number of spikes than a lower intensity pixel.

\textbf{SNN Architecture:} It consists of spiking neurons and interconnecting synapses. 
Here, we consider the architecture in Fig.~\ref{Fig:SNN_TopologyAccuracy}(a), since it has demonstrated robustness when performing different variants of STDP rules for unsupervised learning \cite{Ref:Diehl_STDPmnist_FNCOM15}.
It consists of input, excitatory, and inhibitory layers.
The input layer contains an input image, where every pixel is connected to all excitatory neurons.
In this manner, each excitatory neuron has to recognize a class in the dataset, and the connecting synapses from the same neuron have to learn the features of the corresponding class. 
The excitatory neurons are connected to inhibitory neurons in a one-to-one connection. 
Each spike from an excitatory neuron triggers the corresponding inhibitory neuron to generate a spike that will be delivered to all excitatory neurons, except for the one from which the inhibitory neuron receives a connection.
This inhibition provides competition among excitatory neurons.
Here, a winner-takes-all (WTA) mechanism is employed.

\begin{figure}[hbtp]
\vspace{-0.1cm}
\centering
\includegraphics[width=\linewidth]{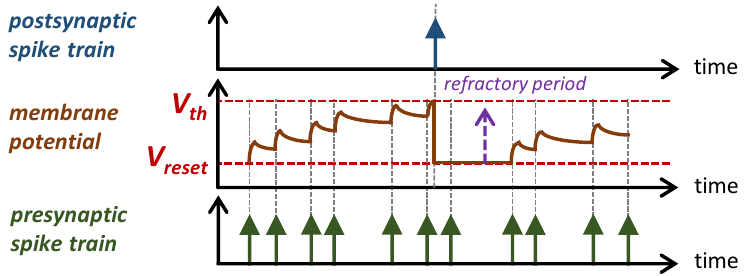}
\vspace{-0.4cm}
\caption{Illustration of the neuronal dynamics of LIF model.}
\label{Fig:LIFillustration}
\end{figure}

\textbf{Neuron Model:} It represents the neuron dynamics and defines neuronal operations.
Here, we consider the Leaky Integrate-and-Fire (LIF) neuron model presented in \cite{Ref:Diehl_STDPmnist_FNCOM15} since it has the lowest computational complexity compared to other biological plausible models \cite{Ref:Izhikevich_CompareModels_TNN04}, which is in-line with our objective. 
Note, LIF has also been widely adopted by the neuromorphic hardware community.
The model describes the dynamics of membrane potential ($V$) as stated in Eq.~\ref{Eq:LIF}. 
\begin{equation}
\begin{split}
\tau_v \frac{dV}{dt} = (E_{rest}-V)+g_e(E_{exc}-V)+g_i(E_{inh}-V)
\label{Eq:LIF}
\end{split}
\end{equation}
Term $\tau_v$ is the membrane time constant. 
$E_{exc}$, $E_{inh}$ are the equilibrium potentials of the excitatory and inhibitory synapses, respectively. 
$E_{rest}$ is the resting membrane potential. 
And, $g_e$, $g_i$ are the conductances of the excitatory and inhibitory synapses, respectively.
The membrane potential $V$ is increased at the occurrence of the incoming spike, otherwise it decays exponentially.
When the membrane potential reaches the threshold potential ($V_{th}$), it generates a spike and goes back to the reset potential ($V_{reset}$).
Afterwards, the neuron is in the refractory period in which it cannot generate spike(s). 
Fig. \ref{Fig:LIFillustration} shows the illustration of the neuronal dynamics of LIF neuron model.

At the system level, we consider adding an adaptive thresholding mechanism to ensure that a neuron does not dominate the spiking activity, and to enable different neurons to recognize different input features, as has been demonstrated in \cite{Ref:Diehl_STDPmnist_FNCOM15}.
Therefore, the membrane threshold is not determined by $V_{th}$ only, rather by the sum of $V_{th}+\theta$, where $\theta$ is increased each time the neuron generates a spike, otherwise the membrane threshold decays exponentially.

\textbf{Synaptic Model and Learning Rule:} 
A synapse is modeled by the conductance change and synaptic weight ($w$), i.e., when a presynaptic spike arrives at a synapse, the conductance is increased by the synaptic weight $w$, otherwise it decays exponentially.
The synaptic weight $w$ is defined by the spike-timing-dependent plasticity (STDP) learning rule, which will be discussed in Section~\ref{Sec:PrelimSTDP}.
The synaptic model is stated as Eq.~\ref{Eq:Synapse}.
\begin{equation}
\begin{split}
\tau_{g_e} \frac{dg_e}{dt} = -g_e \; \; \text{and} \; \; \tau_{g_i} \frac{dg_i}{dt} = -g_i
\label{Eq:Synapse}
\end{split}
\end{equation}
The term $\tau_{g_e}$ denotes the time constant of an excitatory postsynaptic potential, and $\tau_{g_i}$ denotes the time constant of an inhibitory postsynaptic potential.

\subsection{Spike-Timing-Dependent Plasticity (STDP)}
\label{Sec:PrelimSTDP}

\begin{figure}[hbtp]
\vspace{-0.2cm}
\centering
\includegraphics[width=\linewidth]{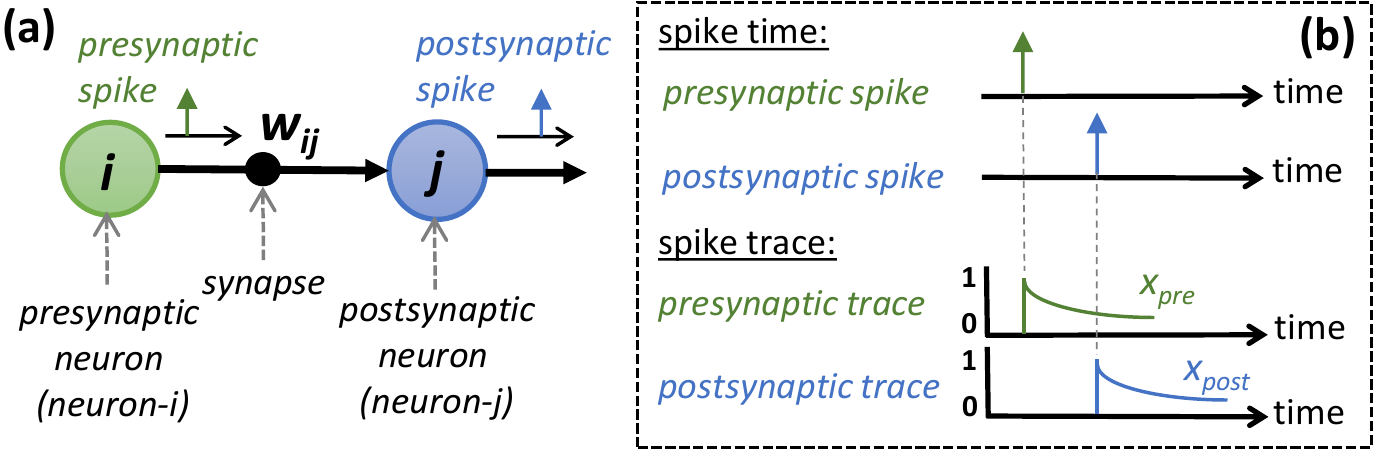}
\vspace{-0.4cm}
\caption{(a) A single synaptic SNN connection. (b) Relation between spike time and spike traces ($x_{pre}$ and $x_{post}$).}
\label{Fig:ConnectSNNnPairSTDP}
\vspace{-0.2cm}
\end{figure}

The synaptic weight change is dependent on the timing correlation between presynaptic and postsynaptic spikes, known as spike-timing-dependent plasticity (STDP) rule \cite{Ref:Bi_STDP_NeuroSci98}.
Although there are several variants of STDP \cite{Ref:Diehl_STDPmnist_FNCOM15}, we consider the general STDP (i.e., pair-wise weight-dependent STDP) as the baseline, since it has been extensively used by previous works \cite{Ref:Diehl_STDPmnist_FNCOM15}\cite{Ref:Hazan_SOMSNN_IJCNN18}\cite{Ref:Hazan_LMSNN_AMAI19}\cite{Ref:Saunders_STDPpatch_IJCNN18}. 
It updates the synaptic weight every presynaptic and postsynaptic spike, based on its temporal correlation with the most recent postsynaptic and presynaptic spike, respectively. 
To improve the simulation speed, the weight changes in STDP are computed using synaptic traces \cite{Ref:Morrison_SynapticTrace_NeCo07}. 
Eq.~\ref{Eq:PairWeightSTDP} is the most common and general form of STDP operation used in literature. 
\begin{equation}
\begin{split}
\Delta w = 
\begin{cases}
-\eta_{pre} x_{post} w^\mu & \text{on} \; \text{presynaptic spike}\\
\eta_{post} x_{pre} (w_{m}-w)^\mu & \text{on} \; \text{postsynaptic spike}
\end{cases}
\label{Eq:PairWeightSTDP}
\end{split}
\end{equation}
$\Delta w$ is the synaptic weight change. $\eta_{pre}$ and $\eta_{post}$ are the learning rate for a presynaptic and postsynaptic spike, respectively. 
$x_{pre}$ and $x_{post}$ are the presynaptic and postsynaptic traces/history, respectively. 
$w_m$ is the maximum weight allowed, $w$ is the previous weight value, and $\mu$ is the weight dependence factor.
Every time a presynaptic spike occurs, $x_{pre}$ is set to 1, otherwise $x_{pre}$ decays exponentially. 
Similar processing is done for the postsynaptic spike using $x_{post}$ (see Fig.~\ref{Fig:ConnectSNNnPairSTDP}). 

\section{Our FSpiNN Framework for Embedded SNNs}
\label{Sec:FSpiNN}

\begin{figure}[hbtp]
\vspace{-0.2cm}
\centering
\includegraphics[width=\linewidth]{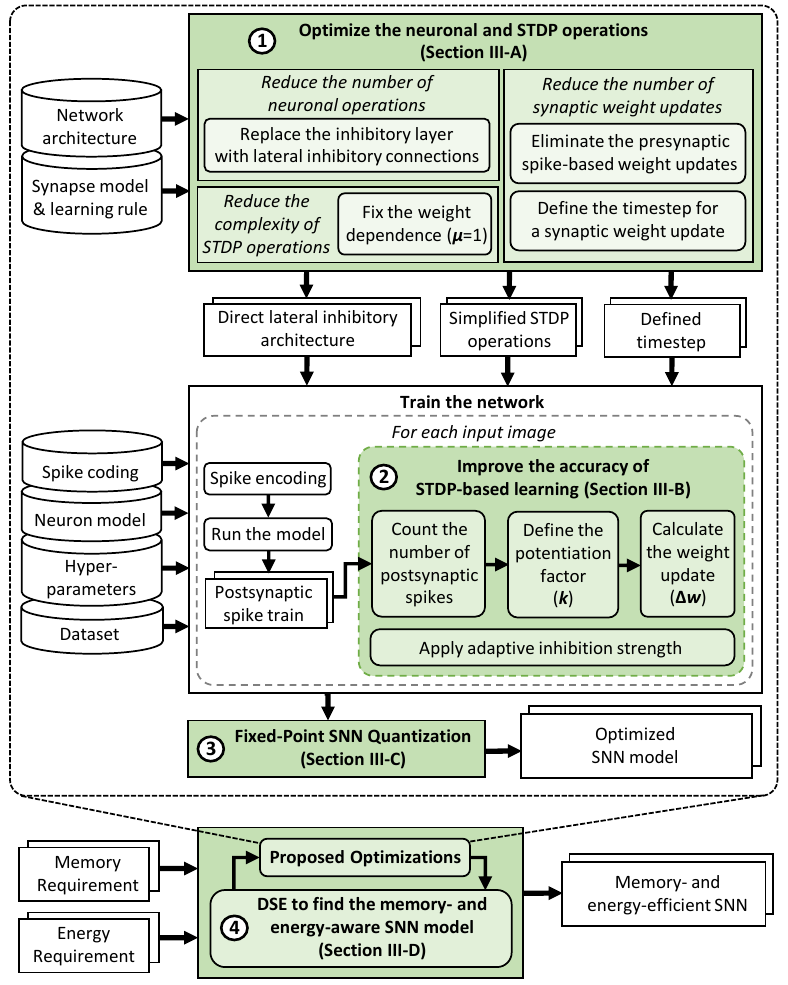}
\vspace{-0.5cm}
\caption{Detailed flow of different steps of our FSpiNN framework. The novel steps are highlighted in green boxes.}
\label{Fig:FSpiNN}
\vspace{-0.1cm}
\end{figure}

We propose FSpiNN, an optimization framework for obtaining memory- and energy-efficient SNNs for both the training and inference while maintaining accuracy.
FSpiNN employs the following key steps. 
The detailed flow of different steps is shown in Fig.~\ref{Fig:FSpiNN}. 

\begin{enumerate} [leftmargin=*]
    \item \textbf{Optimize the processing of neuronal and STDP operations} through the following means (details in \textbf{Section~\ref{Sec:Method_OptSNNcompute}}): 
    \begin{itemize}
        \item Reduce the number of neuronal operations by replacing the inhibitory layer with the direct lateral inhibitory connections. 
        It removes the inhibitory neurons and substitutes the function of spikes from the inhibitory neurons with spikes from the excitatory neurons.
        \item Reduce the number of STDP-based synaptic weight updates by eliminating the presynaptic spike-based weight updates. 
        The updates happen only when the postsynaptic spikes occur, which indicates that the synapses learn the input features effectively. 
        \item Reduce the STDP complexity by fixing the weight dependence factor $\mu$ to 1, hence eliminating the complex exponential calculation. 
    \end{itemize}
    \item \textbf{Improve the accuracy of STDP-based learning} through the following means (details in \textbf{Section~\ref{Sec:Method_ImprovedSTDP}}):
    \begin{itemize}
        \item Timestep-based synaptic weight updates aim to minimize the spurious weight updates that are induced by postsynaptic spikes, thereby ensuring that each update is essential.
        \item Adaptive STDP potentiation factor makes use of the number of postsynaptic spikes to ensure how strong the potentiation should be applied in each weight update. 
        It compensates for the loss of accuracy induced by the STDP simplification. 
        \item Adaptive inhibition strength aims to proportionally provide competition among the excitatory neurons by applying a proper inhibition strength to other neurons.   
        It is derived from an experimental analysis that investigates the accuracy of different inhibition  strength values.
    \end{itemize}
    \item \textbf{Fixed-Point SNN Quantization} (details in \textbf{Section~\ref{Sec:Method_Quantization}}) to further compress the bit-width of SNN parameters. 
    It employs the rounding to the nearest value technique, and explores the trade-off between the accuracy and memory requirements for different quantization levels.
    \item \textbf{A design space exploration algorithm to find the SNN model that fulfills the memory and energy budgets} (details in \textbf{Section~\ref{Sec:Method_DSE}}). It integrates a search algorithm with the proposed optimization to obtain a model that offers a good trade-off in memory, energy, and accuracy.
\end{enumerate}

\vspace{-0.4cm}
\subsection{Optimizing the Computational Requirements of Neuronal and  STDP Operations}
\label{Sec:Method_OptSNNcompute}

\begin{figure}[hbtp]
\centering
\includegraphics[width=\linewidth]{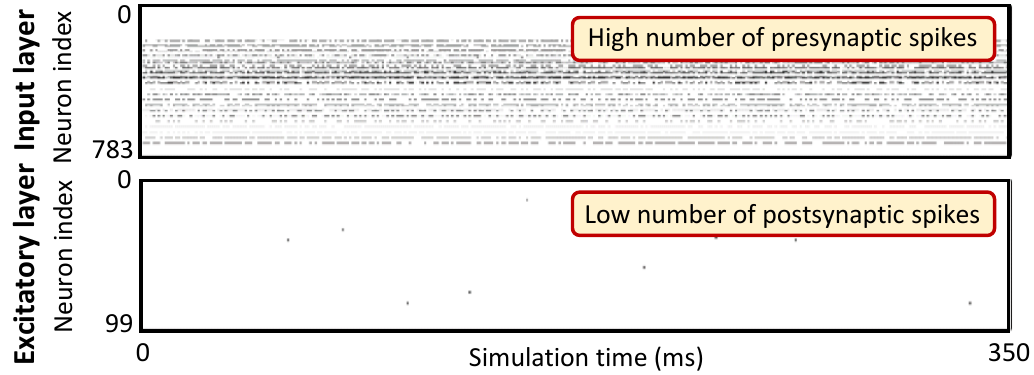}
\vspace{-0.4cm}
\caption{Illustration of the spike trains from the input and excitatory layers. It shows a significant difference between the number of spikes from the input layer (presynaptic spikes) and the excitatory layer (postsynaptic spikes).}
\label{Fig:SpikeObserve}
\vspace{-0.4cm}
\end{figure}

\textbf{Reducing the number of neuronal operations:} 
Our experiments in Fig.~\ref{Fig:SpikeObserve} illustrate that the number of postsynaptic spikes generated from excitatory neurons is less than the presynaptic spikes. 
Therefore, the number of incoming spikes required to trigger an inhibitory neuron to spike is less than the excitatory ones, and the inhibitory neuron typically has a smaller range of active membrane potential (between reset potential $V_{reset}$ and threshold potential $V_{th}$) compared to the excitatory ones. 
This indicates that the inhibitory neurons have different parameters from excitatory ones to be saved in memory. 
Hence, a large number of neurons utilized in the inhibitory layer will consume a considerable amount of memory and energy.
Moreover, each inhibitory neuron needs to process only a small number of incoming spikes to generate the inhibition spike. 
Therefore, the use of inhibitory neurons could be optimized further to reduce the memory and energy requirements.

\begin{figure}[hbtp]
\vspace{-0.2cm}
\centering
\includegraphics[width=\linewidth]{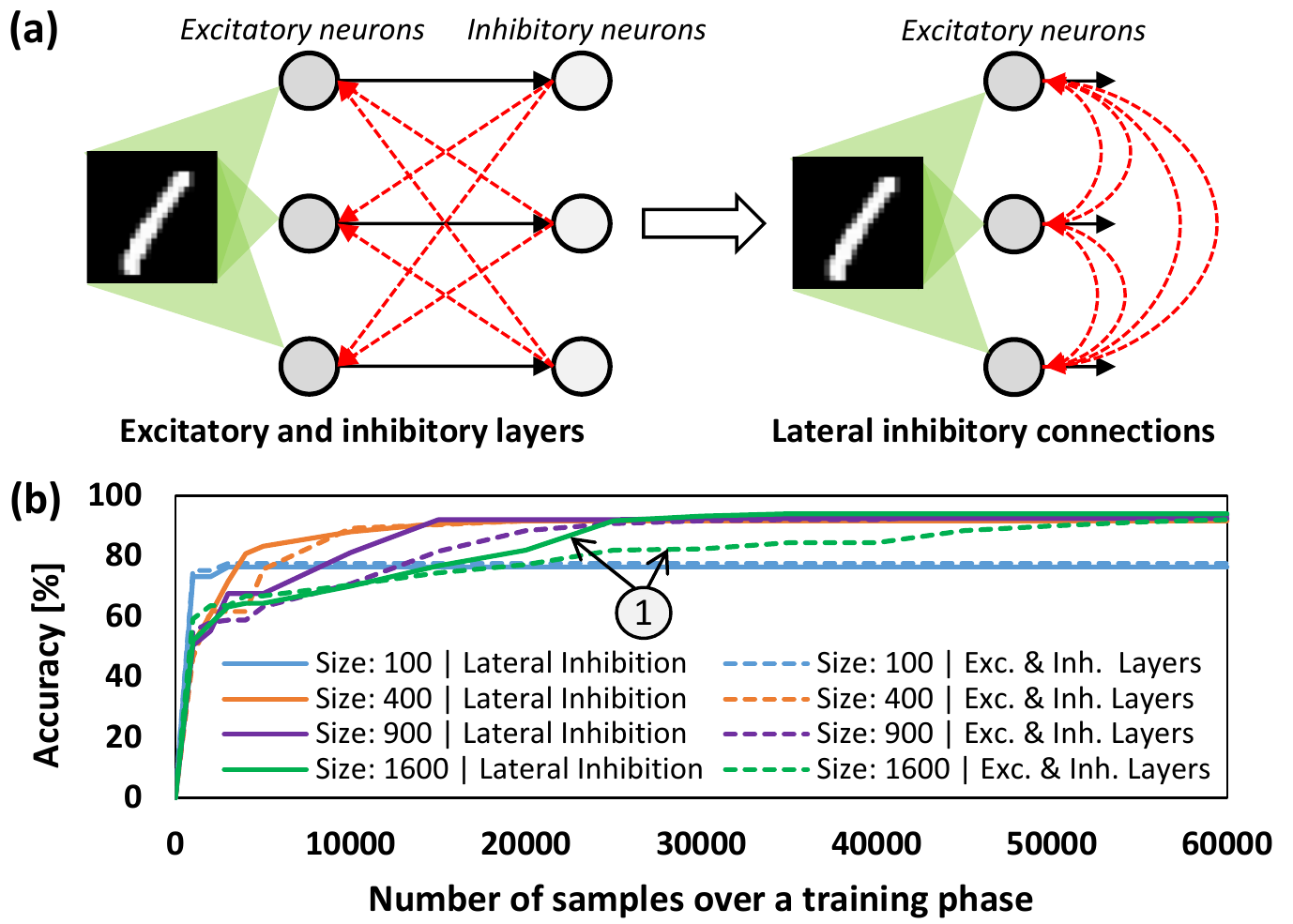}
\vspace{-0.4cm}
\caption{(a) Replacing the inhibitory layer with the direct lateral inhibitory connections (red dashed-lines), through which each excitatory neuron is connected to other excitatory neurons. (b) Impact of employing the direct lateral inhibitory connections on accuracy. This architecture offers comparable accuracy across different sizes of networks (i.e., 100, 400, 900, and 1600 excitatory neurons) as compared to employing the inhibitory layer.}
\label{Fig:ObserveNoInhibLayer}
\end{figure}

Proposed Optimization:
We propose to replace the inhibitory layer with direct lateral inhibitory connections to reduce the number of neurons in the network, thereby curtailing the neuronal operations (as illustrated in Fig.~\ref{Fig:ObserveNoInhibLayer}(a)). 
In this manner, half of the total number of neurons are removed, and the function of spikes from the inhibitory neurons (to provide competition among excitatory neurons through a winner-takes-all mechanism) is substituted by the spikes from the excitatory neurons. 
Our experimental results in Fig.~\ref{Fig:ObserveNoInhibLayer}(b) show that the lateral inhibitory connections have the potential to maintain accuracy, while having less resources than using the inhibitory layer.
For instance, label-\circled{1} in Fig.~\ref{Fig:ObserveNoInhibLayer}(b) indicates that the SNN with a lateral inhibition can achieve a high accuracy faster than the SNN with an inhibitory layer, and then they converge to a comparable accuracy after more samples presented in the training phase.
The reason is that, the lateral inhibition directly conveys spikes from an excitatory neuron to other neurons, hence the number of spikes is typically higher than the ones from the inhibitory layer. 
Therefore, the inhibition is stronger and it results in more diverse feature learning across neurons, thereby achieving high accuracy with a small number of training samples.
This behavior is beneficial, especially when the SNN-based systems have only a small number of training samples.

\textbf{Reducing the number of STDP-based synaptic weight updates:} 
In the unsupervised SNNs, each neuron has to recognize features that belong to a specific class, so that each neuron can generate the highest number of spikes to represent its recognition category. 
To achieve this, the general STDP rule presented in Eq. \ref{Eq:PairWeightSTDP} updates the synaptic weight in every event of a pre- and post-synaptic spike. 
However, previous work \cite{Ref:Srinivasan_EnhPlast_IJCNN17} observed that there are spurious weight updates which may decrease the accuracy of learning.
The spurious updates are observed in two conditions: (i) when the neurons spike unpredictably in the early phase of learning, due to the random weight initialization, and (ii) when a neuron generates spikes for patterns that belong to different classes, but share common features, thereby causing the synapses to learn the overlapped features from different classes. 
Therefore, the STDP-based weight updates that are induced by these pre- and post-synaptic spikes might not learn the input features effectively, and hence decreasing the recognition capability of the neuron and consuming energy.
We exploit this observation in a new way to optimize the SNN computations, while preserving the classification accuracy.

Proposed Optimization:
We propose to eliminate the presynaptic spike-based weight updates to reduce the spurious weight updates that are induced by the presynaptic spikes. 
Therefore, the learning will focus on the condition when postsynaptic spikes happen, which indicates that the connecting synapses effectively learn the input features. 
It also reduces the computational energy as the number of presynaptic spikes is higher than the postsynaptic ones, as shown in Fig.~\ref{Fig:SpikeObserve}.

\vspace{0.4cm}
\textbf{Reducing the STDP Complexity:}
The change in each synaptic weight ($\Delta w$) is updated using an STDP operation that requires complex exponential calculations for the synaptic trace and weight dependence parts (see Eq.~\ref{Eq:PairWeightSTDP}).
We observed that the value of the weight dependence factor ($\mu$) is typically less than 1 \cite{Ref:Srinivasan_EnhPlast_IJCNN17}, which makes it expensive to compute.
Therefore, the use of weight dependence factor could be optimized to achieve further energy-efficiency.

\begin{figure}[hbtp]
\centering
\includegraphics[width=\linewidth]{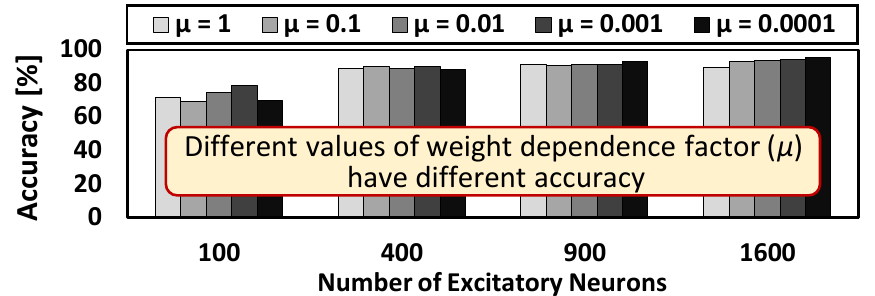}
\vspace{-0.4cm}
\caption{Impact of different values of weight dependence factor $\mu$ on accuracy, across different sizes of networks.}
\label{Fig:ObserveMuWeight}
\end{figure}

Proposed Optimization:
We propose to fix the weight dependence factor $\mu$ to 1, thereby simplifying the computation of STDP operations. 
However, we observed that only fixing the weight dependence factor value may degrade the classification accuracy across different sizes of the network, as shown in Fig.~\ref{Fig:ObserveMuWeight}. 
Therefore, we propose a technique for improving the STDP-based learning (discussed in Section~\ref{Sec:Method_ImprovedSTDP}) to compensate for the loss of this $\mu$ simplification, and to maintain the accuracy.

\subsection{Improving the Accuracy of STDP-based Learning}
\label{Sec:Method_ImprovedSTDP}

We observed that for each input image, at least a single excitatory neuron is expected to recognize the input features and generate the highest number of spikes to represent the recognition of the corresponding class. 
Therefore, information regarding the number of postsynaptic spikes should be leveraged and used to improve the accuracy. 

\begin{figure}[hbtp]
\centering
\includegraphics[width=\linewidth]{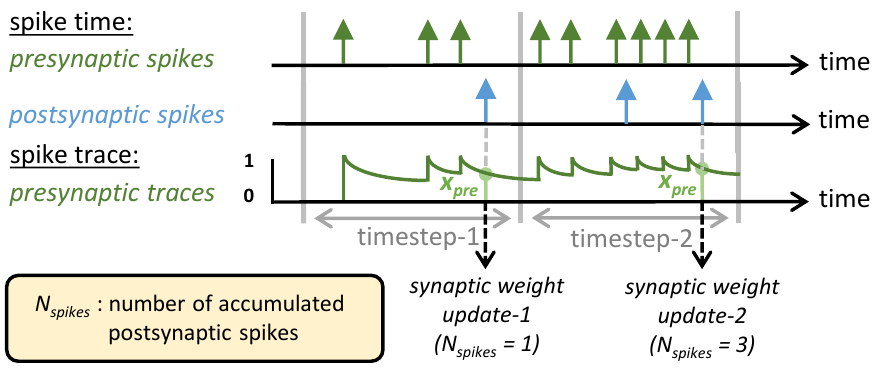}
\vspace{-0.4cm}
\caption{Overview of the timestep, synaptic weight updates, and number of accumulated postsynaptic spikes ($N_{spikes}$) in our proposed technique.}
\label{Fig:STDP_FSpiNN}
\end{figure}

Proposed Solution:
We propose an algorithm to improve the accuracy of STDP-based learning by employing timestep-based synaptic weight updates, and adaptively determining the STDP potentiation factor ($k$) and the inhibition strength. 
Timestep-based synaptic weight updates aim to reduce the spurious weight updates that are induced by the postsynaptic spikes.
Therefore, our technique updates the weight once within a timestep, as long as at least there is a postsynaptic spike (see Fig.~\ref{Fig:STDP_FSpiNN}). 

We also propose an adaptive STDP potentiation factor $k$, which aims at determining how strong the potentiation should be in each weight update, by leveraging the number of postsynaptic spikes.
To do this, our technique accumulates the number of postsynaptic spikes observed from the first time when the spike trains of an input image are presented to the network, until the time when a weight update is performed (denoted as $N_{spikes}$ in Fig.~\ref{Fig:STDP_FSpiNN}).
The number of postsynaptic spikes is used to determine the potentiation factor $k$, as formulated in Eq.~\ref{Eq:STDP_k}.
Term $maxN_{spikes}$ denotes the maximum number of accumulated spikes, and $N_{spikes\_th}$ denotes the number of threshold spikes, which normalizes the value of $maxN_{spikes}$. 
Afterwards, the potentiation factor $k$ is used to compute the synaptic weight change $\Delta w$, as formulated in Eq.~\ref{Eq:ImprovedSTDP}.
The synaptic weight update is conducted for the excitatory neuron that generates the highest number of postsynaptic spikes (i.e., the winning neuron). 
In this manner, the confidence level of learning is expected to increase over time when presenting the spike trains of an input image. 

\begin{equation}
k = \left \lceil \frac{maxN_{spikes}}{N_{spikes\_th}} \right \rceil\\
\label{Eq:STDP_k}
\end{equation}

\begin{equation}
\Delta w = 
k \eta_{post} x_{pre} (w_{m}-w) \; \; \text{on} \; \text{update time} \\
\label{Eq:ImprovedSTDP}
\end{equation}

\begin{figure}[hbtp]
\centering
\includegraphics[width=\linewidth]{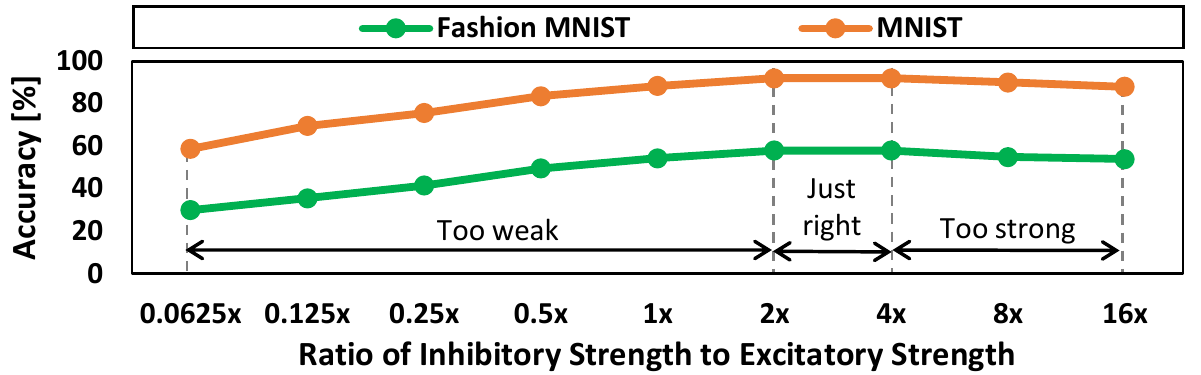}
\vspace{-0.5cm}
\caption{Impact of different ratio values between inhibitory and excitatory strengths when running the MNIST and Fashion MNIST datasets. When the ratio is too weak or too strong, the accuracy is sub-optimal.} 
\label{Fig:ObserveInhibStrength}
\end{figure}

Furthermore, balancing the strength of excitatory and inhibitory synaptic conductance is important as it makes the inhibition neither too strong, nor too weak. 
Too strong inhibition means that once the winning neuron is selected, it strongly prevents other excitatory neurons from firing, thereby dominating the recognition of input features (ineffective competition).
Meanwhile, too weak inhibition means that it does not necessarily provide competition among the excitatory neurons, thereby giving no influence to the overall learning process (no competition).
Previous work in \cite{Ref:Diehl_STDPmnist_FNCOM15} observed that the ratio between the excitatory and inhibitory strengths have an important role to balance the learning process. 
Towards this, we performed an experimental analysis to investigate the accuracy in different inhibition strength conditions and different datasets to justify the generality of the effective ratio conclusion. 
The results are presented in Fig.~\ref{Fig:ObserveInhibStrength}.
Our analysis shows that when the inhibitory strength is too weak or too strong, the accuracy is sub-optimal.
We observed that two comparable accuracy points are obtained using the ratio of 2x-4x.
Therefore, we propose to use an adaptive inhibition strength that provides a proper competition among the excitatory neurons, by applying an inhibition strength equal to 2x-4x of the excitatory strength.

Algorithm~\ref{Alg:ImpSTDPopts} synergistically employs the above-discussed techniques. 
For each excitatory neuron, the algorithm monitors whether a postsynaptic spike is generated. 
If so, the number of the postsynaptic spikes is accumulated, and the corresponding presynaptic traces are recorded.  
Otherwise, no action is required.
When the timestep is reached, the algorithm identifies which excitatory neuron generates the highest number of spikes (the winning neuron).
Once a winning neuron is identified, the connecting synapses to the winning neuron are updated with the synaptic weight change $\Delta w$.

\begin{algorithm}[hbtp]
\caption{\color{black} Pseudo-code for improving the accuracy of STDP-based learning}
\label{Alg:ImpSTDPopts}
\begin{algorithmic}[1]
\renewcommand{\algorithmicrequire}{\textbf{INPUT:}}
\renewcommand{\algorithmicensure}{\textbf{OUTPUT:}}
\REQUIRE \textbf{(1)} Number of training dataset ($D_{train}$); \\ 
\textbf{(2)} Simulation time for an input image ($t_{sim}=350$); \\ 
\textbf{(3)} Timestep ($t_{step}=4$); \\
\textbf{(4)} SNN parameters: number of excitatory neurons ($n_{exc}$), number of synapses-per-neuron ($n_{syn}$), number of accumulated postsynaptic spikes ($N_{spikes}$); \\
\textbf{(5)} STDP parameters: learning rate ($\eta_{post}=0.01$), max. weight value ($w_m=1$), previous weight value ($w$), number of threshold spikes ($N_{spikes\_th}=10$), potentiation factor ($k$);\\ 
\textbf{(6)} Postsynaptic spike event ($spike_{post}$);\\
\ENSURE Synaptic weight update ($\Delta w$); \\
\vspace{0.1cm}
\renewcommand{\algorithmicrequire}{\textbf{BEGIN}}
\renewcommand{\algorithmicensure}{\textbf{END}}
\REQUIRE \hspace{0.1cm} \\   
    \textbf{Initialization}: \\
      \STATE $\Delta w[n_{exc}, n_{syn}] = zeros[n_{exc},n_{syn}]$; \\
      \STATE $N_{spikes}[n_{exc}] = zeros[n_{exc}]$; \\
      \STATE $x_{pre} = zeros[n_{exc},n_{syn}]$; \\
    \textbf{Process}: \\
      \FOR{$(d = 0$ to $(D_{train}-1))$}
        \FOR{$(t = 0$ to $(t_{sim}-1))$}
          \FOR{$(i = 0$ to ($n_{exc}-1))$}
            \IF{$spike_{post}$}
              \STATE $N_{spikes}[i]$ +$=$ $1$;
              \STATE monitor $x_{pre}[i,:]$;
            \ENDIF\\
          \ENDFOR\\
          \IF{$((t \mod t_{step})$ == $0)$}
            \STATE $maxN_{spikes} = max(N_{spikes})$; 
            \STATE $j \leftarrow index(max(N_{spikes}))$;
            \STATE $k = \lceil (maxN_{spikes}/N_{spikes\_th}) \rceil$;
            \STATE $\Delta w[j,:] = k \eta_{post} x_{pre}[j,:] (w_m-w)$; 
          \ENDIF\\
        \ENDFOR\\
      \ENDFOR\\
    \RETURN $\Delta w$; \\
\ENSURE 
\end{algorithmic}
\end{algorithm}
\setlength{\textfloatsep}{6pt}

\subsection{Fixed-Point Quantization for SNNs}
\label{Sec:Method_Quantization}

It is a common practice to perform SNN processing using a single-precision floating-point operation to achieve a high classification accuracy. 
However, floating-point operations typically consume high memory and energy.
To achieve a memory- and energy-efficient SNN processing, it is more convenient to use a fixed-point format for neuronal and STDP operations.
However, quantizing a value implies a reduction of its representation capability, thereby decreasing the accuracy of the networks.
Therefore, the quantization process should consider the trade-off between accuracy and memory requirement, to find the acceptable quantization levels.
In this manner, the users can select the acceptable accuracy and memory to comply with the design specifications.

Towards this, our FSpiNN framework performs exploration to investigate the impact of different quantization levels of SNN parameters (i.e., synaptic weights) to the accuracy, using a rounding to the nearest value technique with the rounding half-up rule.
It approximates the values that are half-way between two representable numbers by rounding them up.
The fixed-point number can be written in $\langle Q_i.Q_f \rangle$ format, with $Q_i$ and $Q_f$ are the integer and fractional part, respectively.
The total number of bits (wordlength $N$) in the fixed-point format consists of the number of bits for the integer part $N_i$ and the fractional part $N_f$ (i.e., $N = N_i+N_f$). 
The precision of the fixed-point format $\epsilon$ is defined as $\epsilon = 2^{-N_f}$ and it is used to define the quantized number $x_q$. 
\begin{equation}
x_q = \left \lfloor x + \frac{\epsilon}{2} \right \rfloor\\
\label{Eq:Quantized}
\end{equation}

\subsection{Design Space Exploration (DSE) Algorithm for the Memory- and Energy-Aware SNN Model}
\label{Sec:Method_DSE}

To provide better applicability in many application scenarios, the proposed optimizations need to fulfill the given memory and energy requirements. 
Towards this, we also propose a DSE algorithm to find an SNN model whose memory and energy (for both the training and inference) are within the given memory and energy budgets, while maintaining the accuracy.
The main idea is to incrementally increase the size of SNN model (i.e., number of excitatory neurons) and evaluate whether the currently investigated model satisfies the memory and energy budgets. 
If so, the DSE will evaluate whether the accuracy is better.
If the accuracy is the same, the DSE will select the smaller model to keep the memory and energy consumption low.
In this manner, our FSpiNN framework can support many applications where the memory and energy are constrained.
The pseudo-code of the algorithm is presented in Algorithm \ref{Alg:DSE}.

\begin{algorithm}[t]
\caption{Pseudo-code for the DSE algorithm}
\label{Alg:DSE}
\begin{algorithmic}[1]
\renewcommand{\algorithmicrequire}{\textbf{INPUT:}}
\renewcommand{\algorithmicensure}{\textbf{OUTPUT:}}
\REQUIRE \textbf{(1)} Memory requirement ($mem$); \\ 
\textbf{(2)} Energy requirement for training ($E_{train}$); \\ 
\textbf{(3)} Energy requirement for inference ($E_{inf}$); \\
\textbf{(4)} SNN model ($model$): number of the excitatory neurons ($model.n_{exc}$), model size ($model.mem$), energy of model for training ($model.E_{train}$), energy of model for inference ($model.E_{inf}$), accuracy of model ($model.acc$); \\
\textbf{(5)} Number of additional excitatory neurons ($n_{add}$); \\
\ENSURE SNN model ($model$); \\
\vspace{0.1cm}
\renewcommand{\algorithmicrequire}{\textbf{BEGIN}}
\renewcommand{\algorithmicensure}{\textbf{END}}
\REQUIRE \hspace{0.1cm} \\   
    \textbf{Initialization}: \\
      \STATE $model.n_{exc} = 0$; \\
      \STATE $model.size = 0$; \\
      \STATE $acc_{saved} = 0$; \\
    \textbf{Process}: \\
      \WHILE{$model.size \leq mem_{req}$}
        \IF{($model.n_{exc} > 0$)}
          \STATE perform \textit{training} using \textit{Algorithm} \ref{Alg:ImpSTDPopts}; \\
          \STATE monitor $model.E_{train}$; \\
          \IF{($model.E_{train} \leq E_{train})$}
            \STATE perform \textit{inference}; \\
            \STATE monitor $model.E_{inf}$ and $model.acc$; \\
            \IF{($model.E_{inf} \leq E_{inf}$) and ($model.acc > acc_{saved}$)}
              \STATE $acc_{saved} = model.acc$; \\
              \STATE save $model$; \\
            \ENDIF \\
          \ENDIF \\
        \ENDIF \\
        \STATE $model.n_{exc} += n_{add}$;
      \ENDWHILE \\
    \RETURN $model$; \\
\ENSURE 
\end{algorithmic} 
\end{algorithm}
\setlength{\textfloatsep}{6pt}

\section{Evaluation Methodology}
\label{Sec:EvalMethod}

\begin{figure}[hbtp]
\centering
\includegraphics[width=\linewidth]{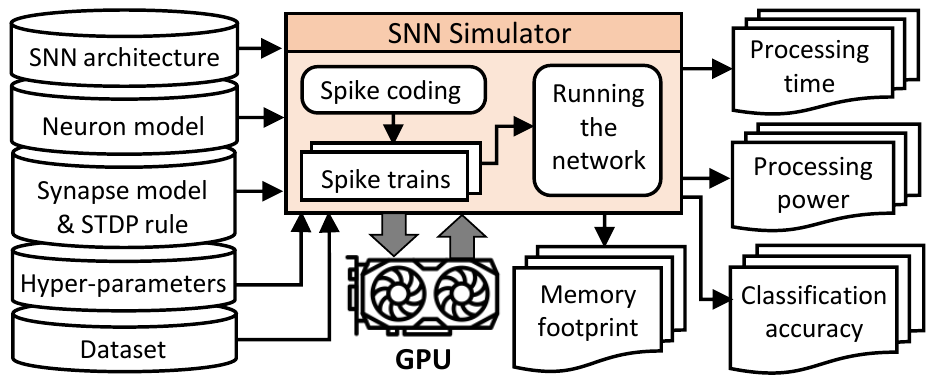}
\vspace{-0.4cm}
\caption{Experimental setup and tools flow.}
\label{Fig:ExpSetup}
\end{figure}

Fig. \ref{Fig:ExpSetup} illustrates the experimental setup with different steps, to evaluate our proposed framework.
We used a Python-based SNN simulator \cite{Ref:Hazan_BindsNET_FNINF18} for evaluating the accuracy of the SNN.  
We run the SNN simulations on three different types of GPUs, namely Nvidia GeForce GTX 1060 \cite{Ref_GTX1060}, GTX 1080 Ti \cite{Ref_GTX1080Ti}, and RTX 2080 Ti \cite{Ref_RTX2080Ti} (the detailed specifications are presented in Table \ref{Table:GPUspecs}), providing a wide range of compute and memory capabilities to show the scalability of our proposed framework.
We selected the GTX 1060 and the GTX 1080 Ti  as representative of the Pascal architecture, which is also used in the embedded GPUs, such as Nvidia Jetson TX2 \cite{Ref_JetsonTX2}. 
We also selected the RTX 2080 Ti as representative of the Turing architecture, to provide variation in compute and memory capabilities for evaluation.  
The same GPU architecture means the same technology and memory hierarchy.  
Therefore, the results obtained from the experiments can be used to estimate the relative energy-efficiency improvement obtained by our FSpiNN framework, as compared to the state-of-the-art works. 

From the simulations, we extracted the size of SNN model which represents the memory footprint.
This information is used to evaluate the memory savings.
To estimate the energy, we adopted the approach of \cite{Ref:Han_DeepCompress_ICLR16}. 
We recorded the start- and end-time of simulation to obtain the processing time, and utilized the \textit{nvidia-smi} utility to report the processing power, which are then used to estimate the energy consumption.

\begin{table}[hbtp]
\vspace{-0.1cm}
\caption{GPU specifications.}
\label{Table:GPUspecs}
\centering
\begin{tabular}{|c|c|c|c|}
\hline 
\textbf{Specifications} & \textbf{GTX 1060} & \textbf{GTX 1080 Ti} & \textbf{RTX 2080 Ti} \\
\hline
\hline
Architecture & Pascal & Pascal & Turing \\
\hline
CUDA cores & 1280 & 3584 & 4352 \\
\hline
Memory & 6GB GDDR5 & 11GB GDDR5X & 11GB GDDR6 \\
\hline
Interface width & 192-bit & 352-bit & 352-bit \\
\hline
Bandwidth & 8Gbps & 11Gbps & 14 Gbps\\
\hline
Power & 120W & 250W & 250W\\
\hline
\end{tabular}
\end{table}

\textbf{Datasets:} 
We used the MNIST \cite{Ref:Lecun_MNIST_IEEE98} and Fashion MNIST \cite{Ref:Xiao_FMNIST_arXiv17} datasets, as they are widely used for evaluating the accuracy of SNNs \cite{Ref:Tavanaei_DLSNN_Neunet18}. 
MNIST represents a simple dataset, while Fashion MNIST represents a more complex dataset \cite{Ref:She_ParallelSpikeSim_DATE19}.
Each dataset has 60,000 images for training and 10,000 images for test, each having a dimension of 28x28 pixels.

\textbf{Input Encoding:} 
Every pixel of an image from the dataset is converted into a Poisson-distributed spike train whose firing rate is proportional to the intensity of the pixel. 
A higher intensity pixel is converted into a higher number of spikes than a lower intensity pixel.
The spike train from each pixel is presented to the network for 350 ms duration.

\textbf{Classification:} 
In the training, the synaptic weight updates are performed without label information as it is unsupervised learning. 
Therefore, an additional mechanism is required to categorize the excitatory neurons for classification. 
The neurons are categorized based on their highest response to different classes over one presentation of the training set (1x epoch of training).
Here, the labels are used to assign each neuron with a specific class. 
Afterwards, the response of the class-assigned neurons is used to measure the accuracy.

\textbf{Comparisons:} 
We compared our proposed framework with two state-of-the-art designs, i.e., the general pair-wise weight dependence STDP-based SNN (baseline) \cite{Ref:Diehl_STDPmnist_FNCOM15}, and the enhanced self-learning STDP-based SNN (SL-STDP) \cite{Ref:Srinivasan_EnhPlast_IJCNN17}.
The sizes of networks considered in the evaluation are the networks with a different number of excitatory neurons: 100, 400, 900, 1600, 2500, 3600, and 4900. 
For conciseness, we refer them to as Net100, Net400, Net900, Net1600, Net2500, Net3600, and Net4900, respectively.
To provide fair comparisons, we recreated the baseline \cite{Ref:Diehl_STDPmnist_FNCOM15} and the SL-STDP \cite{Ref:Srinivasan_EnhPlast_IJCNN17}, and then simulated them using the same SNN simulator \cite{Ref:Hazan_BindsNET_FNINF18}.
We also used the same approach for obtaining the memory footprint and the energy. 
That is, we extracted the size of SNN model from simulation to evaluate the memory footprint, and we used the \textit{nvidia-smi} utility to report the power and recorded the simulation time, which are then used to estimate the energy. 
We also kept the hyper-parameter values the same for different sizes of networks. 
In particular, we used 1x epoch of training because the network will be trained with a full training set once. 
Moreover, an SNN model trained with 1x epoch of training is adopted by a wide-range of SNN community and considered as a completely trained network \cite{Ref:Hazan_SOMSNN_IJCNN18}\cite{Ref:Srinivasan_EnhPlast_IJCNN17}\cite{Ref:Saunders_LCSNN_NeuNet19}\cite{Ref:Saunders_STDPpatch_IJCNN18}.
 
\section{Results and Discussions}
\label{Sec:Results}

\subsection{Maintaining the Classification Accuracy}
\label{Sec:Results_Accuracy}

\begin{figure}[hbtp]
\centering
\includegraphics[width=\linewidth]{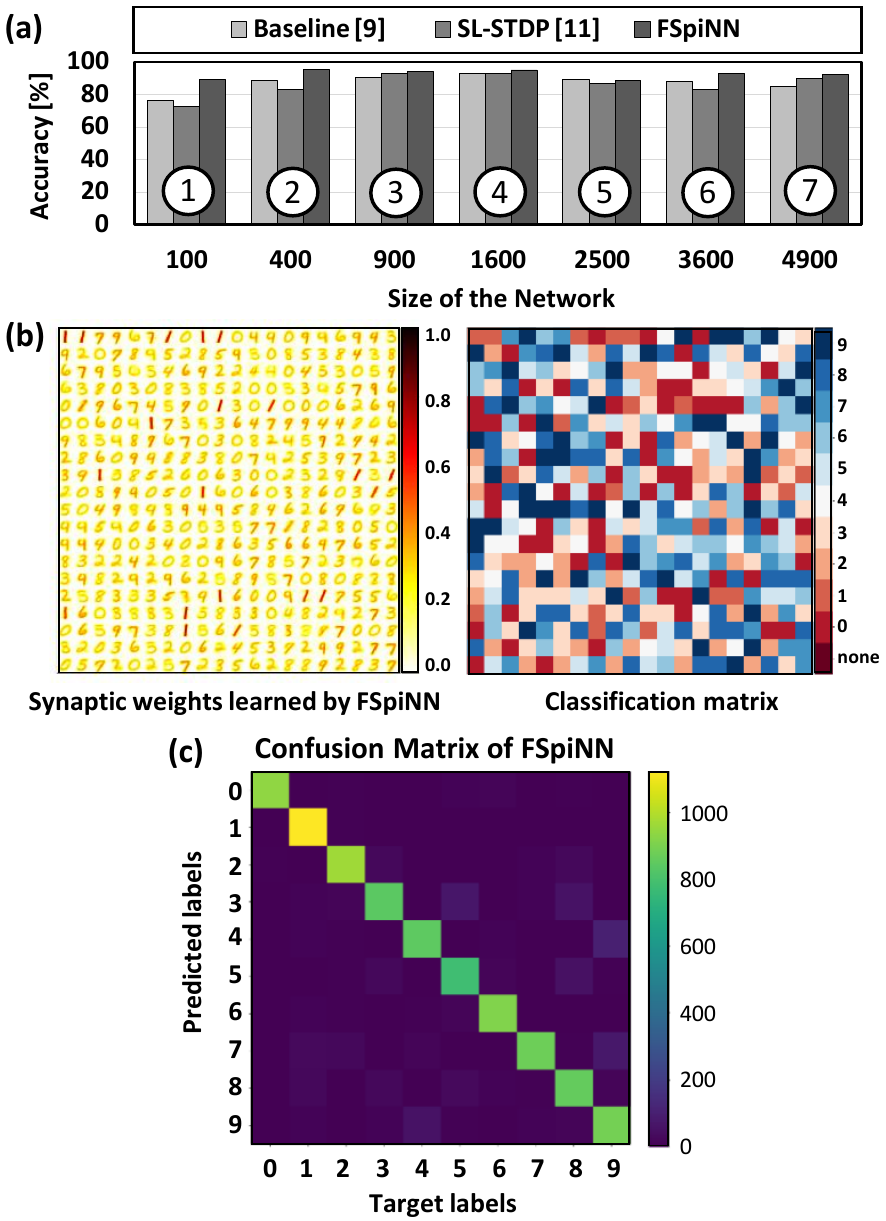}
\caption{(a) Comparisons of accuracy for MNIST dataset in different sizes of networks: Net100, Net400, Net900, Net1600, Net2500, Net3600, and Net4900.
(b) Synaptic weights learned by FSpiNN and its classification matrix. 
(c) Confusion matrix in inference phase for Net400.}
\label{Fig:Result_Accuracy_MNIST}
\end{figure}

\textbf{Results for the MNIST Dataset:} 
Fig.~\ref{Fig:Result_Accuracy_MNIST}(a) shows the accuracy after 1x epoch of training for MNIST. 
It shows that our FSpiNN maintains (and even improves in certain cases) the accuracy across different sizes of networks as compared to other designs. 
Following are the detailed accuracy improvements achieved by FSpiNN: 
\begin{itemize}[leftmargin=*]
    \item Label-\circled{1}: In Net100, FSpiNN achieves 13.2\% improvement with 89.2\% accuracy. 
    \item Label-\circled{2}: In Net400, FSpiNN achieves 7.2\% improvement with 95.6\% accuracy.
    \item Label-\circled{3}: In Net900, FSpiNN achieves 2.4\% improvement with 94.4\% accuracy.
    \item Label-\circled{4}: In Net1600, FSpiNN achieves 2.2\% improvement with 95.2\% accuracy.
    \item Label-\circled{5}: In Net2500, FSpiNN achieves 0.8\% improvement with 90\% accuracy.
    \item Label-\circled{6}: In Net3600, FSpiNN achieves 4.8\% improvement with 92.8\% accuracy.
    \item Label-\circled{7}: In Net4900, FSpiNN achieves 2.4\% improvement with 92.4\% accuracy.
\end{itemize}
These results indicate that a larger network is harder to train.
For instance, the accuracy achieved in Net100 and Net900 are 89.2\% and 94.4{\%} respectively, but the accuracy improvements in Net100 and Net900 are 13.2\% and 2.4\% respectively.
The reason is that, a larger network has more synapses to train for effectively learning the input features, thereby requiring more careful training (e.g., hyper-parameter tuning).
This condition may cause the accuracy of the larger networks lower than the smaller ones in certain cases.
For instance, the accuracy achieved in Net4900 is 92.4{\%}, which is lower than the accuracy in Net900 (i.e., 94.4{\%}).
Furthermore, Fig.~\ref{Fig:Result_Accuracy_MNIST}(b) shows the synaptic weights and its classification matrix, and Fig.~\ref{Fig:Result_Accuracy_MNIST}(c) shows the confusion matrix for Net400. 
These results show the common confusions, such as when identifying between digits 3 and 8, 4 and 9, etc.
The reason is that, the connecting synapses from the same neuron learn the common features (shape) from these classes. Hence, the same neuron generates the highest number of spikes for different classes, thereby resulting in more frequent false classifications.

\begin{figure}[t]
\centering
\includegraphics[width=\linewidth]{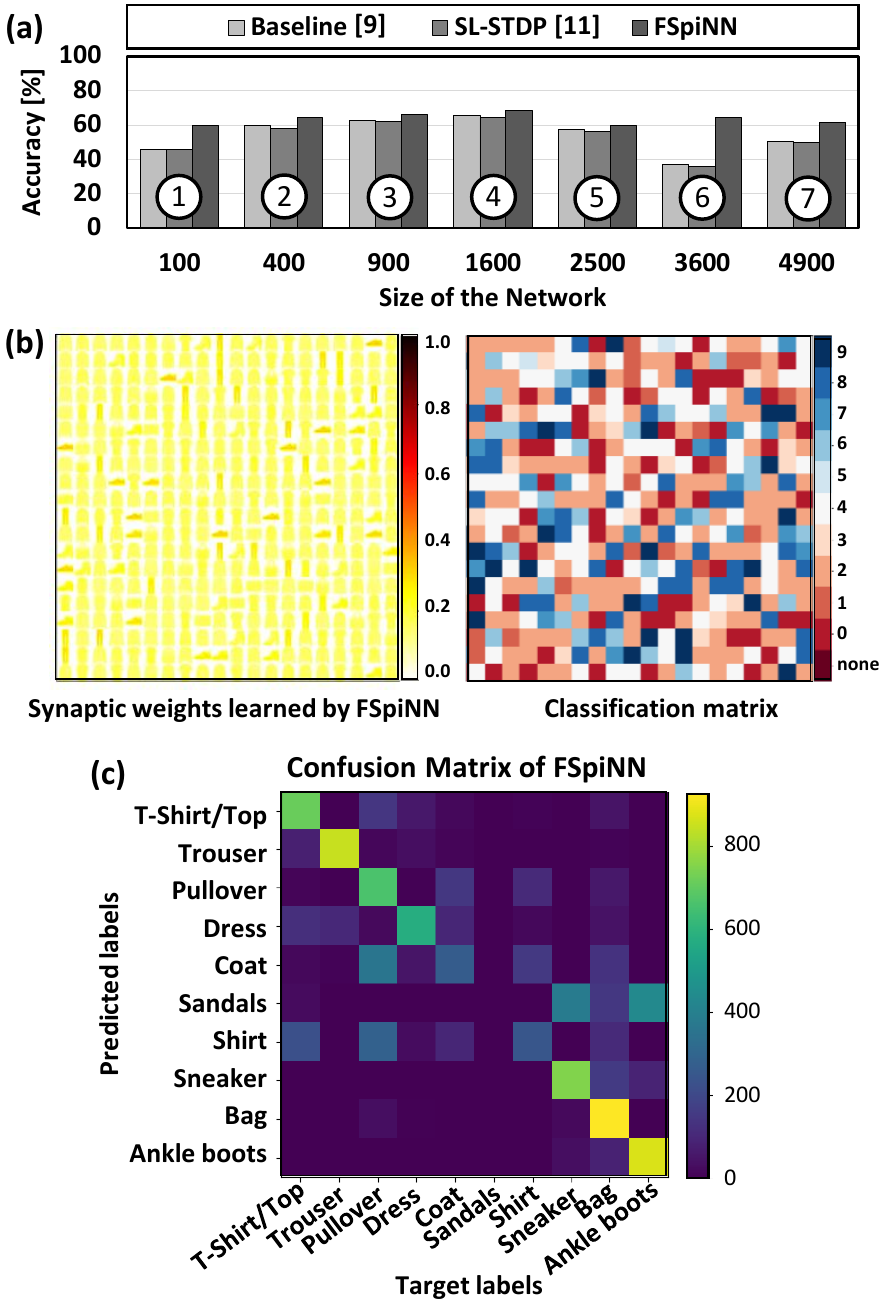}
\caption{(a) Comparisons of accuracy for Fashion MNIST dataset in different sizes of networks: Net100, Net400, Net900, Net1600, Net2500, Net3600, and Net4900.
(b) Synaptic weights learned by FSpiNN and its classification matrix. 
(c) Confusion matrix in inference phase for Net400.}
\label{Fig:Result_Accuracy_FMNIST}
\end{figure}

\smallskip
\textbf{Results for the Fashion MNIST Dataset:} 
Fig.~\ref{Fig:Result_Accuracy_FMNIST}(a) shows the accuracy after 1x epoch of training for Fashion MNIST. 
It shows that our FSpiNN still maintains (and even improves in certain cases) the accuracy across different sizes of networks as compared to other designs. 
Following are the detailed accuracy improvements achieved by FSpiNN: 
\begin{itemize}[leftmargin=*]
    \item Label-\circled{1}: In Net100, FSpiNN achieves 14.2\% improvement with 60.2\% accuracy. 
    \item Label-\circled{2}: In Net400, FSpiNN achieves 5.2\% improvement with 64.8\% accuracy.
    \item Label-\circled{3}: In Net900, FSpiNN achieves 3.6\% improvement with 66\% accuracy.
    \item Label-\circled{4}: In Net1600, FSpiNN achieves 3.5\% improvement with 68.8\% accuracy.
    \item Label-\circled{5}: In Net2500, FSpiNN achieves 3\% improvement with 60.6\% accuracy.
    \item Label-\circled{6}: In Net3600, FSpiNN achieves 27\% improvement with 64.4\% accuracy.
    \item Label-\circled{7}: In Net4900, FSpiNN achieves 11\% improvement with 61.6\% accuracy.
\end{itemize}

\begin{figure}[hbtp]
\centering
\includegraphics[width=\linewidth]{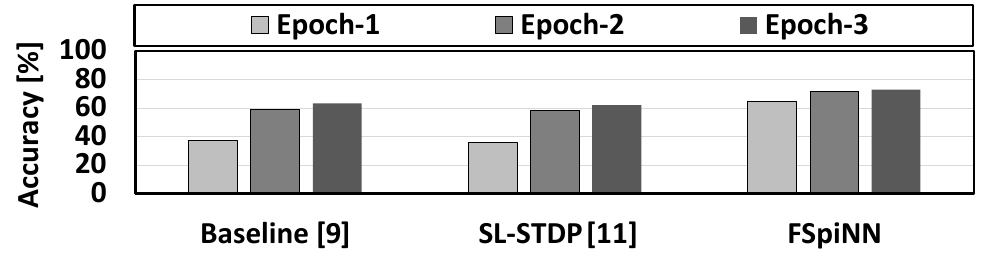}
\caption{Results of accuracy after 3x epochs of training for Net3600 when running the Fashion MNIST.}
\label{Fig:Result_Accuracy_3ep_FMNIST}
\end{figure}

Here, we observed the same trend as observed in MNIST. 
A larger network has the potential to achieve higher accuracy because more neurons are available for recognizing more feature variations. 
This trend is shown in Fig. \ref{Fig:Result_Accuracy_FMNIST}(a) for Net100-Net1600 and Net3600-Net4900. 
At the same time, a larger network is harder to train because more synapses have to effectively learn input features.  
Therefore, a larger network may achieve lower accuracy than the smaller ones in cases where the synapses are not effectively trained. 
This trend is shown in Fig. \ref{Fig:Result_Accuracy_FMNIST}(a) for Net1600-Net3600.
The reason is that, in our experiments, we kept the same hyper-parameter tuning across different sizes of networks, and only performed 1x epoch of training. 
Therefore, the accuracy of a larger network could still be improved through more effective hyper-parameter tuning (e.g., more training epochs), as suggested from Fig. \ref{Fig:Result_Accuracy_3ep_FMNIST}.
The results in Fig. \ref{Fig:Result_Accuracy_3ep_FMNIST} indicate that employing multi-epoch training can increase the accuracy, since the same features in the training set are learned multiple times by the network. 
The accuracy improvement in the earlier epoch is typically higher than in the later ones, thereby only relying on multi-epoch training may incur high energy consumption, without gaining significant accuracy improvement in the end. 
To address this, our FSpiNN employs the adaptive potentiation factor and inhibition strength, which increase the confidence of learning over time in the training.
The results also show that our FSpiNN achieves the highest accuracy across different epochs as compared to state-of-the-art designs. 
Moreover, FSpiNN with 1x training epoch achieves higher accuracy than state-of-the-art designs with 3x training epochs.  
These results show the effectiveness of the learning algorithm in FSpiNN.

Furthermore, Fig.~\ref{Fig:Result_Accuracy_FMNIST}(b) shows the synaptic weights and its classification matrix, and Fig.~\ref{Fig:Result_Accuracy_FMNIST}(c) shows the confusion matrix for Net400. 
These results show the common confusions, such as when identifying between pullover, coat, and shirt, as well as sandals, sneaker, and ankle boots.
The reason is that, the connecting synapses from the same neuron learn the common features (shape) from these classes. 
Hence, the same neuron generates the highest number of spikes for different classes, thereby resulting in more frequent false classifications.

These experimental results also indicate that the input images with more overlapping features are harder to classify. 
Therefore, in general, the classification accuracy achieved in MNIST is higher than Fashion MNIST, since MNIST has relatively simpler features than Fashion MNIST.
However, our FSpiNN can still achieve better accuracy in Fashion MNIST across different sizes of networks, outperforming state-of-the-art designs. 
The maintained accuracy achieved by our FSpiNN comes from the improved STDP-based learning, which reduces the spurious weight updates, and employs the effective STDP potentiation and inhibition strength.

\subsection{Impact of Employing the Fixed-Point Quantization on the Classification Accuracy}
\label{Sec:Results_FixedPoint}

Our framework converts a floating-point (FP32) format to a fixed-point format, and conducts exploration to study the impact of different quantization levels on the accuracy.  

\begin{figure}[hbtp]
\centering
\includegraphics[width=\linewidth]{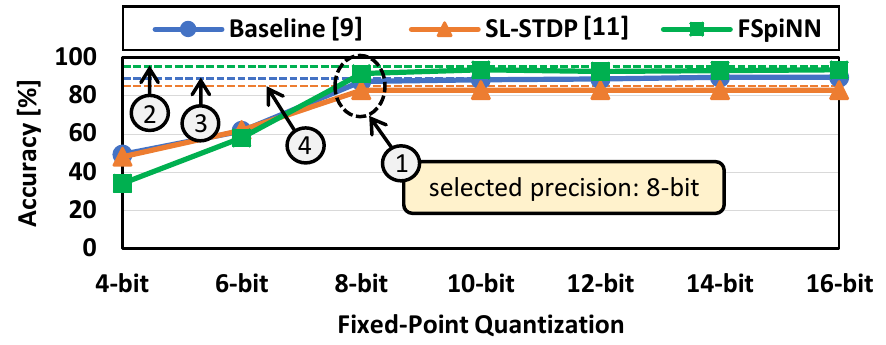}
\caption{Accuracy vs. quantization for MNIST dataset for Net400. 
}
\label{Fig:Result_AccuVFixedP_MNIST}
\end{figure}

\textbf{Results for the MNIST Dataset:} 
Label-\circled{1} in Fig. \ref{Fig:Result_AccuVFixedP_MNIST} shows that the FSpiNN achieves better accuracy than the baseline and the SL-STDP, when the minimum bit-width of quantization is 8 bits.
The reason is that, the 8-bit (or more) format in the FSpiNN provides sufficient levels of weight values to modulate the input spikes from MNIST images, and induce each neuron to recognize a specific digit class. 
In 8-bit precision, our FSpiNN achieves 91.6\% accuracy, while the baseline and the SL-STDP achieve 87.6\% and 82\%, respectively.
It indicates that the accuracy achieved by the FSpiNN 8-bit is slightly less than the FSpiNN FP32 (pointed by the label-\circled{2}), but still higher than the baseline and the SL-STDP with FP32 precision (pointed by the label-\circled{3} and label-\circled{4}, respectively). 
Therefore, the FSpiNN 8-bit offers no accuracy loss with a reduced bit-width for MNIST.

\begin{figure}[hbtp]
\centering
\includegraphics[width=\linewidth]{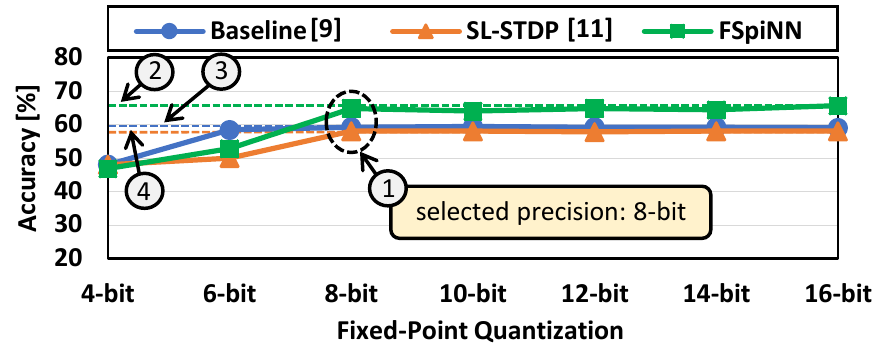}
\caption{Accuracy vs. quantization for Fashion MNIST dataset for Net400. 
}
\label{Fig:Result_AccuVFixedP_FMNIST}
\end{figure}

\smallskip
\textbf{Results for the Fashion MNIST Dataset:} 
Label-\circled{1} in Fig. \ref{Fig:Result_AccuVFixedP_FMNIST} shows that FSpiNN achieves better accuracy than the baseline and the SL-STDP, when the minimum bit-width of quantization is 8 bits.
The reason is that, the 8-bit (or more) format in the FSpiNN provides sufficient levels of weight values to modulate the input spikes from Fashion MNIST images, and induce each neuron to recognize a specific fashion class.
In 8-bit precision, our FSpiNN achieves 64.8\% of accuracy, while the baseline and the SL-STDP achieve 59.2\% and 58\%, respectively.
It indicates that the accuracy achieved by the FSpiNN 8-bit is comparable to the FSpiNN FP32 (pointed by the label-\circled{2}), and it is higher than the baseline and the SL-STDP with FP32 precision (pointed by the label-\circled{3} and label-\circled{4}, respectively).
Therefore, the FSpiNN 8-bit offers no accuracy loss with a reduced bit-width for Fashion MNIST.

These experimental results also show that, for both MNIST and Fashion MNIST datasets, the quantization levels with less than 8-bit precision do not provide sufficient unique information for distinguishing features of different classes in the input images. 
This condition reduces the efficacy of STDP learning of the synapses and recognition capability of the neurons, thereby leading to low classification accuracy.
Furthermore, a reduced bit-width is beneficial since it leads to a reduced memory requirement and energy consumption, which will be discussed in Section~\ref{Sec:Results_Memory} and Section~\ref{Sec:Results_Energy}.
Note, the users can select the quantization level based on the trade-off consideration in the design specifications (e.g., accuracy, memory, and power/energy budget). 

\vspace{-0.1cm}
\subsection{Reducing the Memory Requirements}
\label{Sec:Results_Memory}

\begin{figure}[hbtp]
\centering
\includegraphics[width=\linewidth]{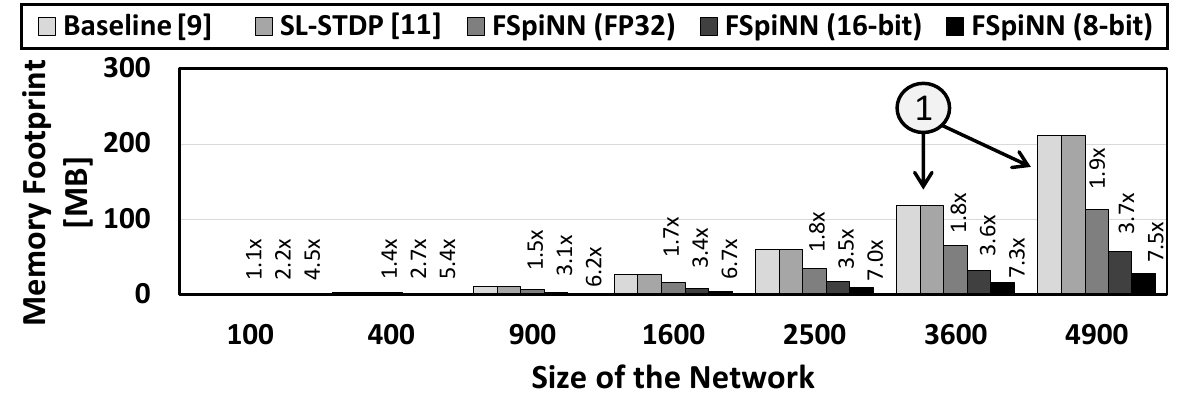}
\caption{Memory requirements for different sizes of networks (i.e., Net100, Net400, Net900, Net1600, Net2500, Net3600, and Net4900) and different quantization levels (i.e., FP32/without quantization, 16-bit, and 8-bit).}
\label{Fig:Result_MemorySaving}
\end{figure}

Fig.~\ref{Fig:Result_MemorySaving} shows the memory requirements of different designs across different sizes of networks for both the training and inference phases.
Label-\circled{1} shows that the Net3600 and Net4900 that employ the baseline or the SL-STDP techniques, consume more than 100MB, thereby making them difficult to be deployed on embedded systems.
On the other hand, our FSpiNN without quantization (FP32) achieves 1.8x and 1.9x memory savings as compared to the baseline, for Net3600 and Net4900, respectively.
The reason is that the FSpiNN FP32 removes the inhibitory neurons completely, thereby avoiding their parameters to be saved in the memory. 
After applying quantization, the memory requirement is reduced even more. 
The FSpiNN 16-bit achieves about 3.6x and 3.7x memory savings, while the FSpiNN 8-bit achieves about 7.3x and 7.5x memory savings, when compared to the baseline for Net3600 and Net4900, respectively.
Fig.~\ref{Fig:Result_MemorySaving} also shows that the FSpiNN 8-bit consumes about 0.16MB-28MB for Net100-Net4900, thereby making the networks easier to be deployed on embedded systems.
Furthermore, if we consider the accuracy that the quantized designs can achieve, we can select the FSpiNN design that offers a good trade-off between high accuracy and acceptable memory footprint.

\vspace{-0.4cm}
\subsection{Energy-Efficiency Improvements}
\label{Sec:Results_Energy}

\begin{figure*}[hbtp]
\centering
\includegraphics[width=\linewidth]{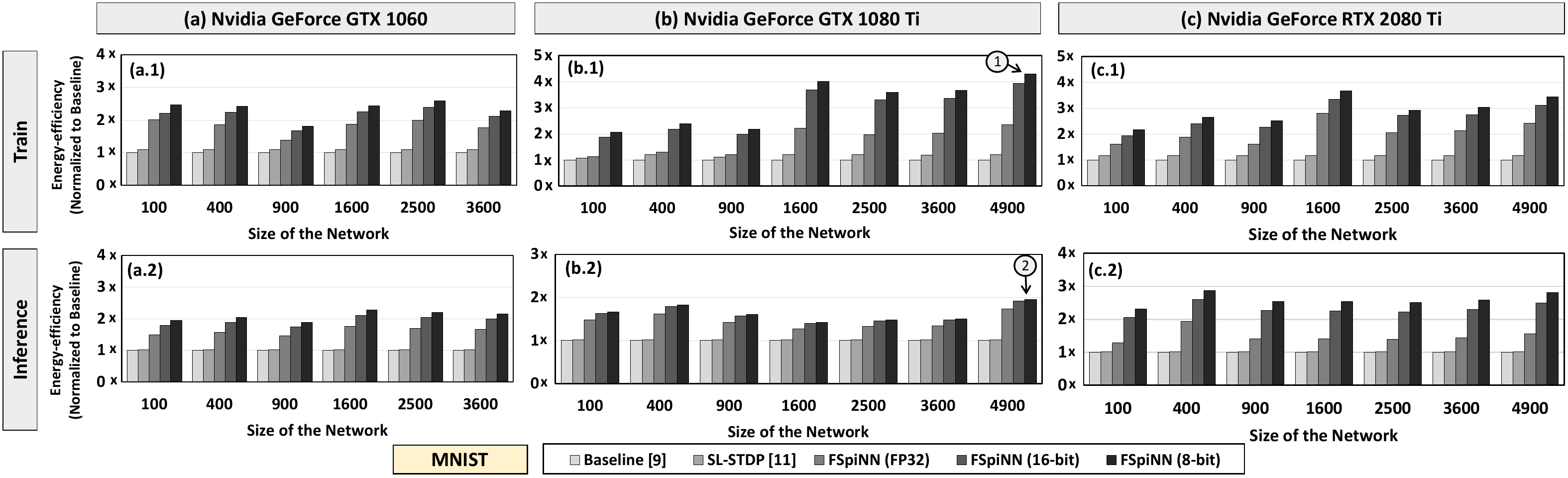}
\caption{Results of energy-efficiency (normalized to the baseline) for training and inference on MNIST, while considering different sizes of networks (i.e., Net100, Net400, Net900, Net1600, Net2500, Net3600, and Net4900), different quantization levels (i.e., FP32, 16-bit, and 8-bit), and different types of GPUs: Nvidia GeForce (a) GTX 1060, (b) GTX 1080 Ti, and (c) RTX 2080 Ti. Due to the limited memory, the GTX 1060 can only run Net100-Net3600.}
\label{Fig_Result_EnergyEff_MNIST_GPUs}
\end{figure*}

\begin{figure*}[hbtp]
\centering
\includegraphics[width=\linewidth]{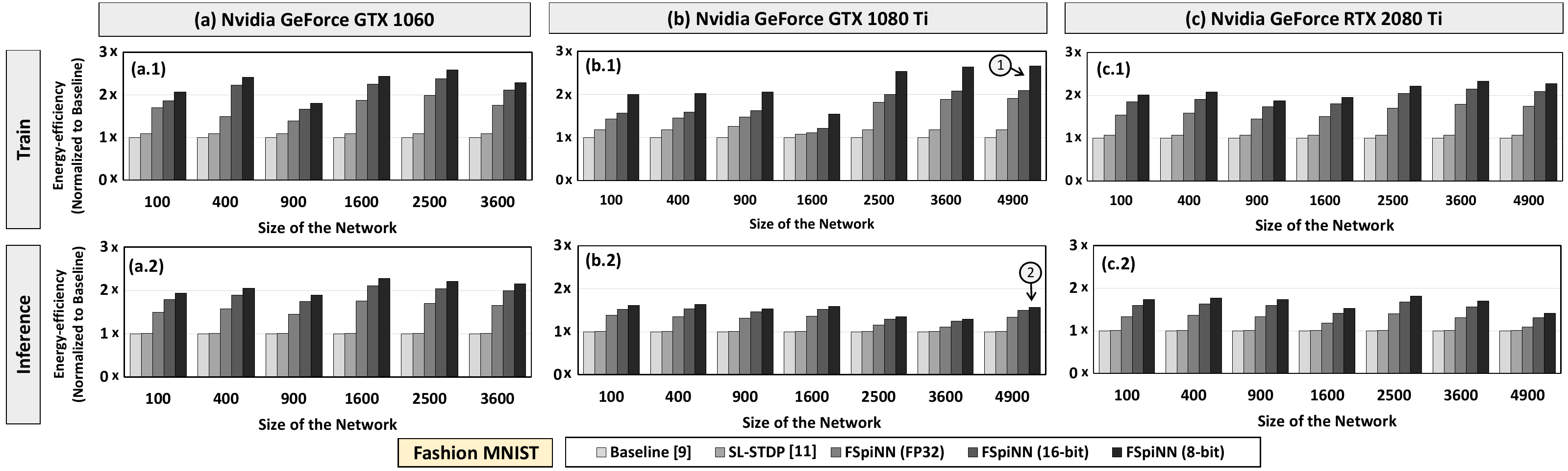}
\caption{Results of energy-efficiency (normalized to the baseline) for training and inference on Fashion MNIST, while considering different sizes of networks (i.e., Net100, Net400, Net900, Net1600, Net2500, Net3600, and Net4900), different quantization levels (i.e., FP32, 16-bit, and 8-bit), and different types of GPUs: Nvidia GeForce (a) GTX 1060, (b) GTX 1080 Ti, and (c) RTX 2080 Ti. Due to the limited memory, the GTX 1060 can only run Net100-Net3600.}
\label{Fig_Result_EnergyEff_FMNIST_GPUs}
\end{figure*}

Fig.~\ref{Fig_Result_EnergyEff_MNIST_GPUs} and Fig.~\ref{Fig_Result_EnergyEff_FMNIST_GPUs} illustrate the energy-efficiency across different sizes of networks and different GPUs for MNIST and Fashion MNIST datasets, respectively. 
These figures show that the SL-STDP achieves higher energy-efficiency than the baseline in training phase, and our FSpiNN achieves the highest energy-efficiency among all designs in both training and inference phases.


\textbf{Training:}
The SL-STDP improves the energy-efficiency by 1.1x-1.2x compared to the baseline, across different sizes of networks and GPUs, for both MNIST and Fashion MNIST.
The reason is that, the SL-STDP only employs postsynaptic spike-based weight updates whose number of updates is less than the baseline, which employs pre- and post-synaptic spike-based weight updates.
The FSpiNN FP32 improves the energy-efficiency more than the SL-STDP, that is by 1.1x-2.8x (MNIST) and by 1.1x-1.9x (Fashion MNIST) compared to the baseline.
The reason is that, apart from the elimination of presynaptic spike-based weight updates, the FSpiNN FP32 also eliminates the inhibitory neurons and reduces the STDP complexity.
After applying quantization, the FSpiNN 16-bit and FSpiNN 8-bit improve the energy-efficiency even more than FSpiNN FP32. 
That is, the FSpiNN 16-bit achieves 1.7x-3.9x (MNIST) and 1.2x-2.4x (Fashion MNIST), while FSpiNN 8-bit achieves a 1.8x-4.3x (MNIST) and 1.5x-2.7x (Fashion MNIST), compared to the baseline.

\textbf{Inference:}
The SL-STDP has comparable energy-efficiency compared to the baseline, across different sizes of networks and GPUs, for both MNIST and Fashion MNIST.
The reason is that, the SL-STDP and the baseline have similar computational complexity in the inference phase.
Meanwhile, the FSpiNN FP32 improves the energy-efficiency by 1.3x-1.9x (MNIST) and by 1.1x-1.4x (Fashion MNIST) compared to the baseline. 
The improvements mainly come from the elimination of the inhibitory neurons.
After applying quantization, the FSpiNN 16-bit and FSpiNN 8-bit improve the energy-efficiency even more than FSpiNN FP32. 
That is, the FSpiNN 16-bit achieves 1.4x-2.6x (MNIST) and 1.2x-2.1x (Fashion MNIST), while FSpiNN 8-bit achieves 1.4x-2.9x (MNIST) and 1.3x-2.3x (Fashion MNIST), compared to the baseline. 

Furthermore, if we consider the classification accuracy and memory footprint that the quantized designs can achieve, we can select the FSpiNN design that offers a good trade-off in the accuracy, memory, and energy-efficiency.
For instance, the FSpiNN 8-bit achieves energy-efficiency improvements by 4.3x (MNIST) and by 2.7x (Fashion MNIST) in training (see labels-\circled{1} in Fig.~\ref{Fig_Result_EnergyEff_MNIST_GPUs} and Fig.~\ref{Fig_Result_EnergyEff_FMNIST_GPUs}), and by 2x (MNIST) and by 1.6x (Fashion MNIST) in inference (see labels-\circled{2} in Fig.~\ref{Fig_Result_EnergyEff_MNIST_GPUs} and Fig.~\ref{Fig_Result_EnergyEff_FMNIST_GPUs}), compared to the baseline in Net4900, while obtaining 7.5x memory saving with an accuracy of $\sim$92\% for MNIST and $\sim$61\% for Fashion MNIST.
The experimental results in Fig.~\ref{Fig_Result_EnergyEff_MNIST_GPUs} and Fig.~\ref{Fig_Result_EnergyEff_FMNIST_GPUs} also 
suggest that our FSpiNN framework is scalable for different sizes of networks and can be used for other systems where different types of GPUs are deployed, such as embedded systems with embedded GPUs.

Note that this work is not about justifying SNNs over deep neural networks (DNNs). 
Rather, we consider what necessary optimizations are required if the SNNs make it to a real-world system  following an increasing trend of the neuromorphic computing, due to their benefits in energy-efficient spike-based computations and unsupervised learning. 
Moreover, there is a substantial difference in the underlying learning mechanism between the SNNs (with the unsupervised learning) and the DNNs (with the supervised learning), thus we cannot directly compare the accuracy of the unsupervised SNNs with the supervised DNNs. 
Previous work \cite{Ref:Du_Compare_MICRO15} has observed that the accuracy of the DNNs (with the supervised  back-propagation algorithm) is generally higher than the SNNs (with the unsupervised STDP algorithm), because the unsupervised STDP algorithm does not have labels when updating the weights, hence it is less effective than the supervised ones.
Furthermore, in the SNN community, many different optimization aspects are explored, and they have the potential to be incorporated into our FSpiNN framework. 
For instance, the works in \cite{Ref:Mohemmed_IJCNN11} and \cite{Ref:Russell_TNN10} focus on generating precise spike sequences like the real-world observation. 
They target a different optimization purpose compared to the one targeted by our FSpiNN framework. 
However, they can still be incorporated in the FSpiNNs' optimization flow for generating precise spike sequences.
This illustrates the flexibility of our FSpiNN for integration with other optimization techniques.

\section{Conclusion}
\label{Sec:Conclusion}
In this paper, we proposed a novel FSpiNN framework that synergistically employs different techniques to reduce the memory footprint and to improve the energy-efficiency of SNNs, while maintaining their accuracy. 
Experimental results illustrate the benefits and efficiency of the proposed framework, compared to the state-of-the-art designs, across different sizes of networks and different datasets (MNIST and Fashion MNIST).
For instance, in a network with 4900 excitatory neurons, our FSpiNN achieves 7.5x memory saving and 3.5x energy-efficiency improvement on average for training and by 1.8x on average for inference, with no accuracy loss.
In short, our proposed framework enables efficient embedded SNN implementations for the next-generation smart embedded systems.
 





\section*{Acknowledgment}
The authors acknowledge the scholarship granted by the Indonesia Endowment Fund for Education (IEFE/LPDP), Ministry of Finance, Republic of Indonesia.

\ifCLASSOPTIONcaptionsoff
  \newpage
\fi



%
\bibliographystyle{IEEEtran}
\bibliography{bibliography}


%

\vspace{-1cm}
\begin{IEEEbiography}[{\includegraphics[width=1in,height=1.25in,clip,keepaspectratio]{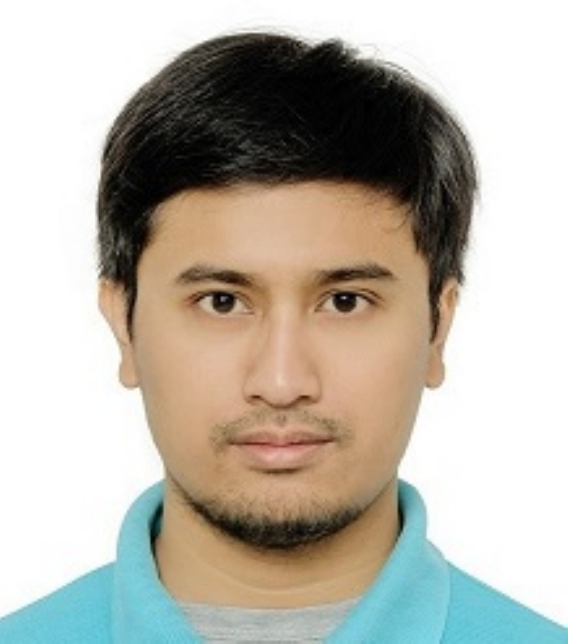}}]
{Rachmad Vidya Wicaksana Putra}
is a research assistant and Ph.D. student at Computer Architecture and Robust Energy-Efficient Technologies (CARE-Tech), Institute of Computer Engineering, Technische Universität Wien (TU Wien), Austria. He received B.Sc. on Electrical Engineering in 2012 and M.Sc. on Electronics in 2015 with distinction, both from Bandung Institute of Technology (ITB), Indonesia. He was a teaching assistant at Electrical Engineering Department, School of Electrical Engineering and Informatics ITB in 2012-2017 and also a research assistant at Microelectronics Center ITB in 2014-2017. He is a recipient of the Indonesian Endowment Fund for Education (IEFE/LPDP) Scholarship from Ministry of Finance, Republic of Indonesia. His research interests mainly include computer architecture, VLSI design, system-on-chip, brain-inspired and neuromorphic computing, energy-efficient computing, and electronic design automation.
\end{IEEEbiography}

\vspace{-1cm}
\begin{IEEEbiography}[{\includegraphics[width=1in,height=1.25in,clip,keepaspectratio]{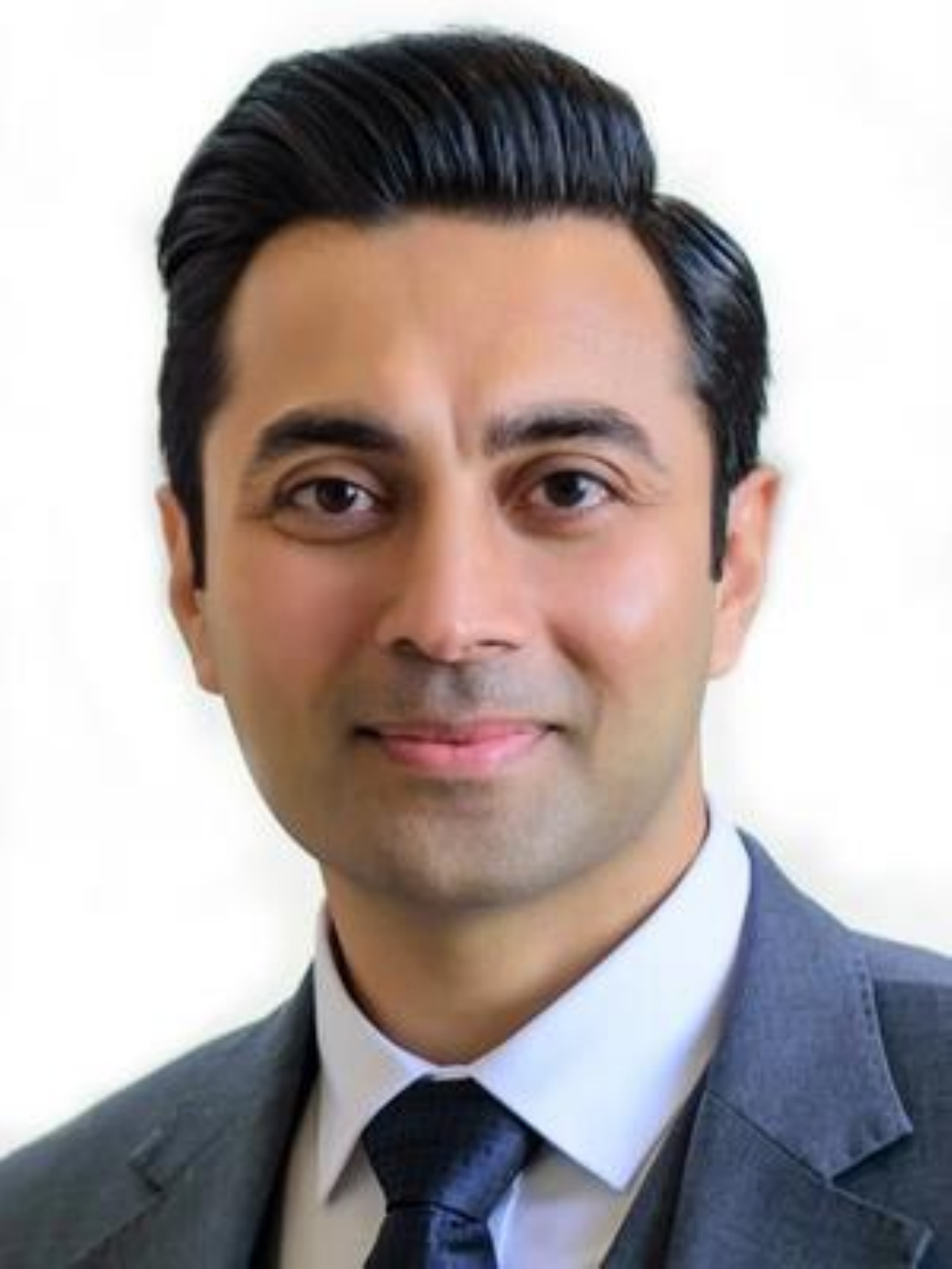}}]
{Muhammad Shafique}(M'11 - SM'16) received the Ph.D. degree in computer science from the Karlsruhe Institute of Technology (KIT), Germany, in 2011. Afterwards, he established and led a highly recognized research group at KIT for several years as well as conducted impactful R\&D activities in Pakistan. In Oct.2016, he joined the Institute of Computer Engineering at the Faculty of Informatics, Technische Universität Wien (TU Wien), Vienna, Austria as a Full Professor of Computer Architecture and Robust, Energy-Efficient Technologies. Since Sep.2020, he is with the Division of Engineering, New York University Abu Dhabi (NYU AD), United Arab Emirates.

His research interests are in brain-inspired computing, AI \& machine learning hardware and system-level design, energy-efficient systems, robust computing, hardware security, emerging technologies, FPGAs, MPSoCs, and embedded systems. His research has a special focus on cross-layer analysis, modeling, design, and optimization of computing and memory systems. The researched technologies and tools are deployed in application use cases from Internet-of-Things (IoT), smart Cyber-Physical Systems (CPS), and ICT for Development (ICT4D) domains.

Dr. Shafique has given several Keynotes, Invited Talks, and Tutorials, as well as organized many special sessions at premier venues. He has served as the PC Chair, Track Chair, and PC member for several prestigious IEEE/ACM conferences. 
Dr. Shafique holds one U.S. patent has (co-)authored 6 Books, 10+ Book Chapters, and over 200 papers in premier journals and conferences. He received the 2015 ACM/SIGDA Outstanding New Faculty Award, AI 2000 Chip Technology Most Influential Scholar Award in 2020, six gold medals, and several best paper awards and nominations at prestigious conferences. \end{IEEEbiography}




\end{document}